\documentclass[letterpaper]{article}

\usepackage{amsmath}
\usepackage{natbib,alifeconf}  
\usepackage{caption,subcaption}
\usepackage{hyperref}
\usepackage{etoolbox}

\usepackage[T1]{fontenc}
\newtoggle{blindsubmission}

\togglefalse{blindsubmission}

\title{Biomaker CA: a Biome Maker project using Cellular Automata}

\iftoggle{blindsubmission}{
\author {Anonymous Authors}
}{
\author {Ettore Randazzo,$^{1}$ \and Alexander Mordvintsev$^{1}$\\
\mbox{} \\
$^1$ Google Research \\
\{etr, moralex\}@google.com}
}

\begin{document}
\maketitle

\begin{abstract}

We introduce Biomaker CA: a Biome Maker project using Cellular Automata (CA). In Biomaker CA, morphogenesis is a first class citizen and small seeds need to grow into plant-like organisms to survive in a nutrient starved environment and eventually reproduce with variation so that a biome survives for long timelines. We simulate complex biomes by means of CA rules in 2D grids and parallelize all of its computation on GPUs through the Python JAX framework. We show how this project allows for several different kinds of environments and laws of 'physics', alongside different model architectures and mutation strategies. We further analyze some configurations to show how plant agents can grow, survive, reproduce, and evolve, forming stable and unstable biomes. We then demonstrate how one can meta-evolve models to survive in a harsh environment either through end-to-end meta-evolution or by a more surgical and efficient approach, called Petri dish meta-evolution. Finally, we show how to perform interactive evolution, where the user decides how to evolve a plant model interactively and then deploys it in a larger environment. We open source Biomaker CA at: \href{https://tinyurl.com/2x8yu34s}{https://tinyurl.com/2x8yu34s}.

\end{abstract}

\begin{figure*}
  \includegraphics[width=\textwidth]{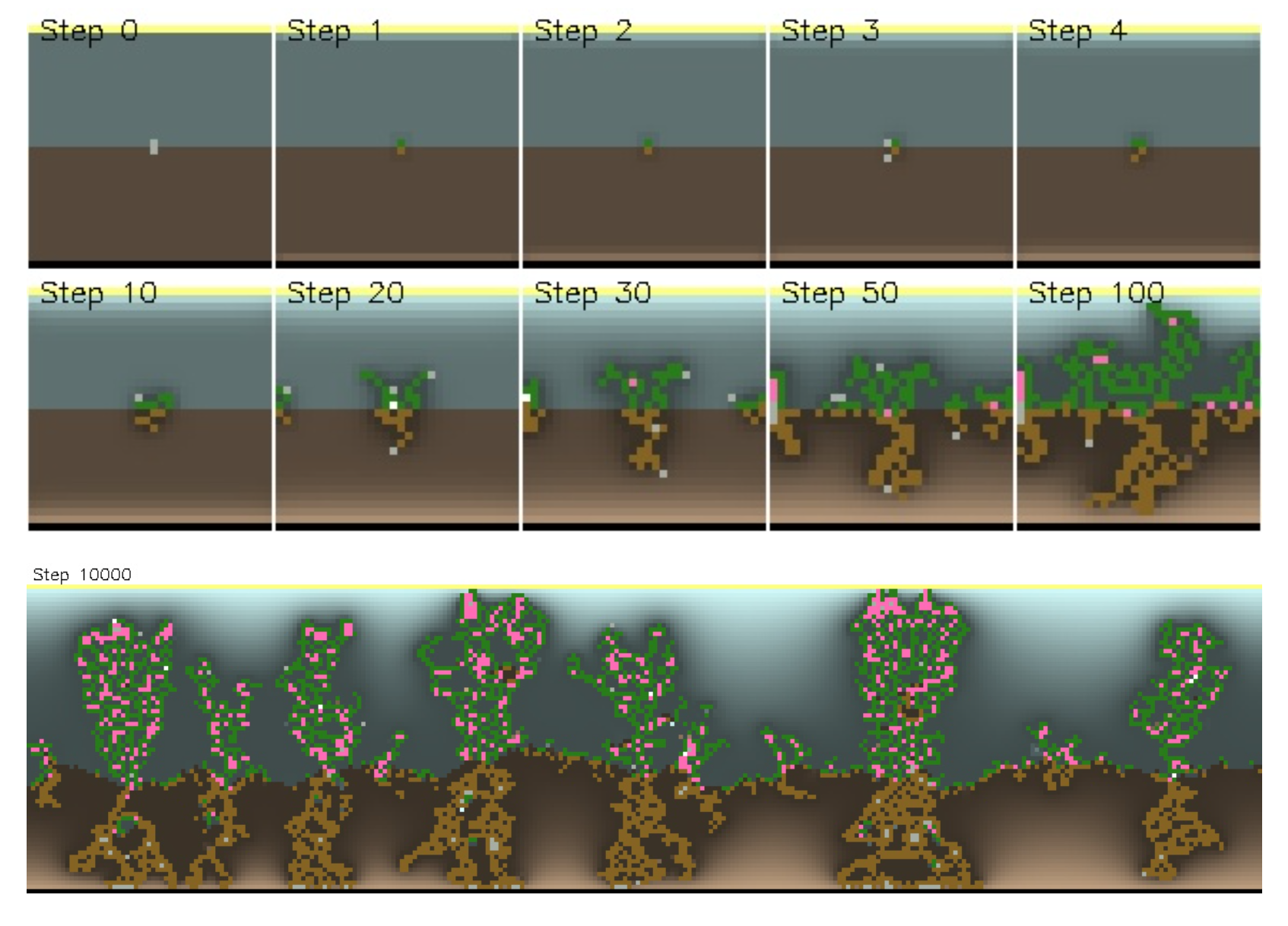}
  \caption{Example runs of the "persistence" configuration. Above: we use a tiny variant of it, showing the initial steps of the simulation. At step 0, there is only a seed of unspecialized cells. At step 1, agents have specialized to leaf and root. At step 3, both agents perform a "spawn" operation. Future steps show a typical unfolding and reproduction. The brightness of the ground and air represents how many nutrients are present. The environment starts scarce with nutrients but nutrients get quickly diffused from its top/bottom sides. Below: in a wider variant of the configuration, we show an example result after 10000 steps. Several plants grow very tall and starve competing neighbors out of resources. The amount of earth in the environment changes constantly, due to agents' growth and death.}
  \label{fig:hero_figure}
\end{figure*}

\section{Introduction}

\textit{Biomaker CA} is a \textit{Biome Maker} project using Cellular Automata (CA). With Biomaker CA, different kinds of environments can be designed where agent seeds can be placed in order to observe a plant-based life simulation in action. In Biomaker CA, morphogenesis is a first-class citizen. Plants grow, try to survive, reproduce with variation and eventually die, all of this while having to learn to sustain a complex metabolism. The resulting lifeforms generate constantly changing biomes where organisms have to compete for resources. Biomaker CA runs entirely on GPU and is implemented in Python using JAX \citep{jax2018github} to allow for both interactive and large scale experiments.

The aims of this project are multiple. We expect Biomaker CA to be useful in the research of growing complexity (or complexification), open endedness, evolvability, artificial life and ecology. Before we describe the system in details, we believe it is beneficial to discuss the current status of some of the fields highlighted in order to understand where and how Biomaker CA can be of help.

\subsection{Complexification and Open endedness}
In the field of Artificial Life (ALife), one of the main goals is to figure out how to design increasingly complex lifeforms (in vivo or in silico) such that they are able to either survive in (increasingly) complex environments, or express specific capabilities once deploying in a virtual or physical environment. The question of how to achieve arbitrary complexification (the growing of complexity) up to the point of solving any given task, is still unsolved. It is nonetheless undeniable how finding a solution to this problem would be relevant to many other areas of study, such as machine learning and artificial intelligence, mathematics, and philosophy.

Achieving complexification is usually interpreted differently depending on which field of study pursues this goal. In the Deep Learning, Reinforcement Learning and Evolutionary Strategies fields, researchers are most interested in figuring out how to solve a specific set of problems and evaluate the performance of their agents/models with some loss or fitness functions. While their results are often impressive even in the generality of the found solutions \citep{team2021openended} or how prone they are to transfer learning \citep{liu2023summary,plested2022deep}, it is extremely difficult to afterwards take these models and "evolve them" in order to reuse as many parameters as possible and decrease the training time of a new, more complex version of the same model, or for a different task. It is this kind of complexification that we are most interested in with this line of research.

Not all optimization strategies are however solely focused on maximising a fitness function or minimizing a loss. Researchers have discovered how sometimes simply searching for \textit{novelty} is better than trying to blindly optimize for a task for solving the task of interest \citep{Lehman2008ExploitingOT}. They also showed how often optimizing both for novelty \textit{and} the task fitness is better for several use cases \citep{Pugh2016-ao, Mouret2015IlluminatingSS}. Sometimes, a task is so open ended that it is impossible to define any fitness function, but it nevertheless can show significant signs of complexification. Picbreeder \citep{Secretan2011-sl}, for instance, lets the user explore the possible space of images and lets them pursue the very vague metric of "interestingness".

In ALife, complexification is often observed and discussed when talking about \textit{open endedness}. Open endedness is hard to describe formally, but it intuitively means that an environment demonstrates the capability of endless variation, and therefore complexification, of the agents' behaviors interacting in the environment. Here, endless variation implies complexification because any fixed size of complexity would eventually extinguish the available forms of variation, therefore an unbounded increase in complexity is required. The study of open endedness in this sense subsumes complexification, since it also includes other fundamental questions such as what are the initial conditions of a system and what is the selective pressure that together are sufficient for accomplishing endless complexification. In this field, solving a specific task is generally considered insufficient, and truly open ended environments are expected to be able to adapt to different long-term dynamics and effectively evolve agents into solving unpredictable new tasks. There have been attempts to evaluate the open endedness of a system and under some metrics Geb was the first system to be classified as open ended \citep{channon2006geb}, however, regardless on whether this is actually true, the simplicity of the environment makes it hard to see a path forward for generalizations of Geb to more interesting environments. In \citep{soros2014}, the authors try to define a set of necessary and sufficient conditions for achieving open endedness, but it still remains unclear whether the set is truly necessary or sufficient, and it does not describe a procedure on how to accomplish the conditions reliably.

In some recent approaches to create open ended systems, agents and tasks are coevolved, so that tasks become increasingly complex the more complex agents become \citep{wang2019paired, wang2020enhanced}. Similarly, other researchers aim at evolving autotelic agents: agents that figure out what tasks to explore themselves \citep{schmidhuber2013, forestier2017intrinsically}.

A relevant and widespread approach for achieving open endedness is in setting up some systems where there is a very simple "minimal criterion" for reproduction \citep{soros2014, brant2017as}, such as being alive for long enough, and let complexity arise through agents interacting with the environment. As we will see, in the core experiments of Biomaker CA, the only selective pressure is in agents being capable of creating a flower under certain conditions and trigger a reproduce operation, making this the minimal criterion for in-environment evolution.

\subsection{Evolvability}

One feature that seems to be essential for accomplishing this kind of complexification or open endedness is called \textit{evolvability}. There are many different and nuanced definitions of this term (for instance, see \cite{Wagner1996-rt}, \cite{kirschner1998} or \cite{Pigliucci2008-vx}). One definition that seems to be widely accepted is this: "The capacity to generate heritable, selectable phenotypic variation." By this we mean that an agent is more evolvable if its offsprings are more likely to generate \textit{useful} variation in the phenotypic space. The reason why evolvability is so important can be seen with a counterexample: if at a certain point we'd have very low or zero evolvability on our agents, we wouldn't be able to discover any new behaviours.

The study of how to make agents more evolvable explores many different aspects of model and environment design. Chiefly, it seems uncontroversial that modular models are more evolvable \citep{kirschner1998, Wagner1996-rt, Kashtan2005-ym, Clune2013-yk} and that modules need to be adaptable \citep{kirschner1998}. On a related note, it appears that allowing for redundancy may be beneficial as well \citep{ebner2001}. A sound strategy for getting more evolvable agents appears to be having dynamic fitness landscapes \citep{grefenstette1999, Reisinger2006-xi} and extinction events \citep{lehman2015ex}; it is valuable to highlight how open endedness tends to have dynamic fitness landscapes as requirements, hinting at the close relationship between them and evolvability. Encouraging to search for novelty also seems to help evolvability \citep{lehman2011as, lehman2015ex}. Some research shows that, at least with our current model encodings, crossing over results in the loss of evolvability or successful optimization \citep{lehman2011as, Kramer2010}. Finally, there appears to be some conflict as to whether adaptive mutation rates help \citep{Wagner1996-rt, grefenstette1999, mengistu2016} or hinder evolvability \citep{glickman2000, Clune2008-zs}. It is our humble hypothesis that \textit{some} versions of adaptive mutations are necessary (in a similar way that nature is able of keeping some traits constant across eras), while others may fall victim to vicious local minima that could hinder survivability and evolvability.

Some researchers have tried to optimize directly for evolvability by evaluating a parent by first creating offspring and checking for its variance \citep{mengistu2016, lehman2011as, gajewski2019}, while, as we will see in the next subsection, most take away teachings from evolvability research to construct open ended environment that push for evolvable agents naturally.

\subsection{Morphogenesis, self replicators and Cellular Automata}

If there is one system that has certainly accomplished open endedness, and has shown astounding levels of complexification, is biological life. It would therefore be ill-advised to not ponder on what biology does that most of our current systems don't do. One striking developmental process that is lacking in most artificial intelligence and artificial life experiments is the one of morphogenesis. In biology, life starts from a single cell, for the most part, and grows into complex organisms through complex indirect encodings. This, however, is hard to design for small scale experiments, and hard to train, resulting (especially in deep learning) into focusing on direct encodings where there is a very close relationship between genotype and phenotype. In Biomaker CA, we want to investigate whether the use of morphogenetic encodings can result in more evolvable and complexifying systems than would otherwise arise. Notably, some ALife experiments have shown morphogenesis in some forms \citep{channon2006geb, spector2007}, but none to our knowledge have focused so much on it as to make it a core requirement of the system to exist and reproduce.

The field of morphogenesis has a tight connection to the one of self-replication. In a certain sense, most of ALife experiments use self-replication, up to the point that \textit{life} in itself can be considered the quest for the most robust self-replication. Of particular interest for this project are Von Neumann's Cellular Automata \citep{neumann1966ca}, some of the first virtual environments for exploring self replication and artificial life. There is a long history of creating virtual environments with CA as their inhabitants – for instance, the old Game of Life \citep{10.2307/24927642} or the recent Lenia \citep{chan2019lenia} – but recently a neural version CA, called Neural Cellular Automata (NCA), have shown very complex behaviors of morphogenesis \citep{mordvintsev2020growing,mordvintsev2022iso,r2023growing} and interactions with one another \citep{randazzo2020self-classifying, randazzo2021adversarial, cavuoti2022}. In \cite{Sinapayen2023}, the author shows how NCA show signs of self replication with variation purely by the noise in the cells states when organisms replicate. In Biomaker CA, we decided to use some variants of NCA to grow complex morphogenetic organisms. As we will see, however, mutation mostly happens at the \textit{parameters} level, and not at the state level, increasing drastically the potential variability, but with it also its possible risks of chaotic drift \citep{randazzo2021sr}.

\subsection{Simulated environments}

Obviously, we are not the first to come up with a simulated environment. In contrast, there are so many different past and ongoing projects of ALife that we have to restrict ourselves into talking about only a few. The following exposition will be aimed at highlighting common patterns and limitations of past experiments, in order to show how Biomaker CA might be of help where others would be less fit.

First off, there are experiments that create entire environments, but then agents are evaluated one at a time \citep{sims1994,miconi2005tr, wang2019paired, wang2020enhanced}. Due to this limitation, there is no complex interaction among agents that can cause a change in the fitness landscape.

There are environments based on CA or particle systems \citep{mordvintsev2022particle,ventrella2023,Schmickl2016-nc} where the rules of CA or particles are hard coded for the entirety of the simulations. In that case, discovering complexifying life takes the form of observing self-sustaining patterns, or even better, self reproducing ones. The original Lenia \citep{chan2019lenia} is a great CA example where several lifeforms are identified, but often each lifeform is determined by a unique set of parameters that represent the laws of physics, making it unlikely to have different lifeforms in the same environment. The search for open endedness in these kinds of environments would entail finding a fixed set of rules/parameters that cause the emergence of unbounded complexity. However, this approach may be computationally intractable at the moment. Two recent different generalizations of Lenia \citep{chan2023largescale,plantec2022flowlenia} allow for different parameters being used in the same environment and include methods for varying the parameters inside the environments through mutations or parameter recombination. The results show a promising direction where each environment shows more variety than what observed in the original work.

Some environments do show interactions among agents in one form or another, and show very complex potential rule sets that evolve during an environment lifetime, like Tierra \citep{ray-approach-to-synthesis-1991}, Avida \citep{Ofria2004-zg} and Geb \citep{channon2006geb}. However, these environments are very simple and it is unclear whether increase in complexity of agents equals increase in interestingness, or whether we can learn how to solve complexification tasks on other environments with these approaches.

In Neural MMO \citep{suarez2019neural}, complex agents with different parameters interact and compete in the same environment. However, evolution happens externally and not in the environment. Moreover, there is no capability of altering one's bodily functions.

Chromaria \citep{soros2014} is an interesting environment where agents interact with one another, but in a sequential manner, where an agent is added to the environment one at a time, and its fitness is largely dependent on the previous placement and parameterization of other agents. Agents can become arguably arbitrarily complex in their expression of coloring, but the simplicity of the environment may make it unsuitable to discovering general trends for complexification.

There is a plethora of environments where agents interact in an environment and reproduce with variation in the environment itself \citep{spector2007, miconi2008, yaeger1995, channon2006geb, ventrella1998genepool, robinsonlife, heinemann2023, chan2023largescale, plantec2022flowlenia}. They all have each their own strengths and we believe that anybody interested in these topics should check them all out. Where we believe Biomaker CA may distinguish itself from them is the level of complex behavior that can be easily accomplished with it, and its focus on morphogenesis as a first class citizen.

As we stated before, some environments make use of some kind of morphogenesis. Geb \citep{channon2006geb} has a developmental phase using L-systems, but the resulting growth has no environmental effects and only affects the intangible complexity of the logic of the agents. Division Blocks \citep{spector2007} grow by dividing their core components, and "reproduce" by severing connections among components. However, the resulting bodies appear to be mostly simple snake-like organisms.

\subsection{Biomaker CA overview}

\begin{figure}[h]
  \includegraphics[width=\columnwidth]{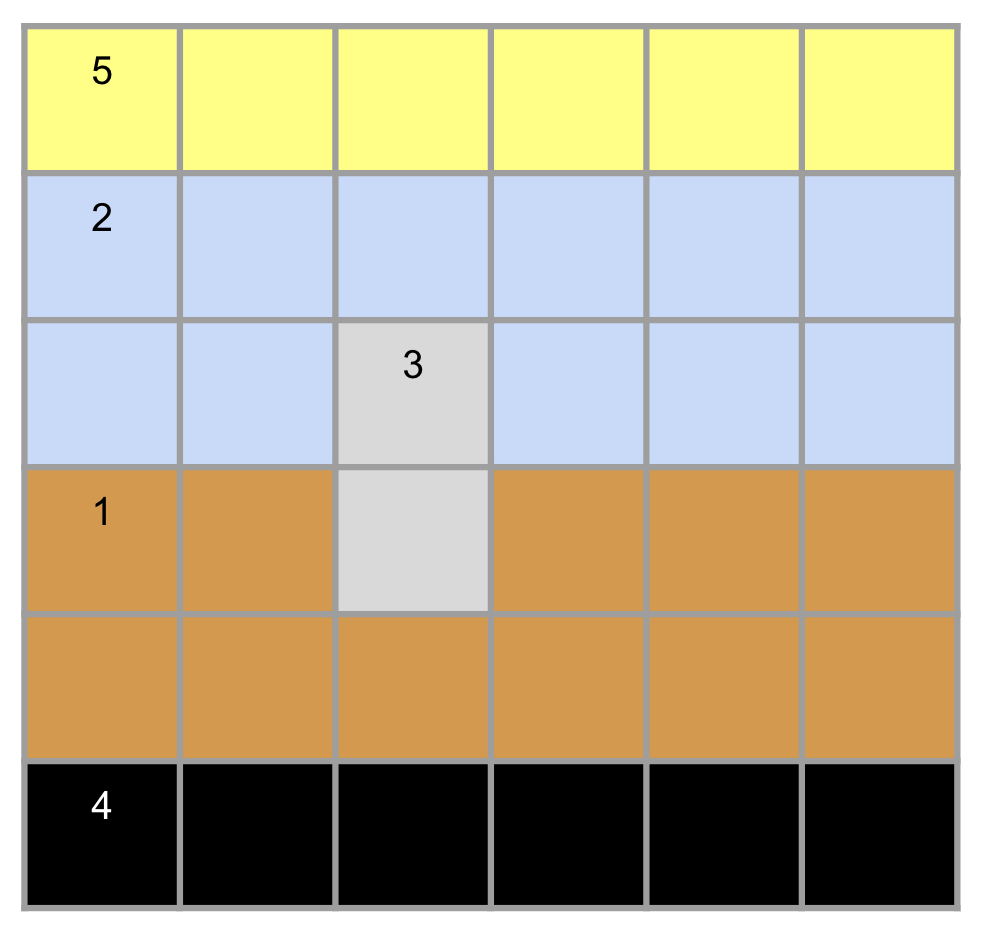}
  \caption{Schematic of the minimal requirements for an environment. These 5 materials need to appear: 1) Earth. 2) Air. 3) Agent; the two gray cells represent a seed. 4) Immovable. 5) Sun.}
  \label{fig:basic_env}
\end{figure}

With the previous projects in mind, we wanted to create a framework that can generate sets of environments where very complex lifeforms can be designed and/or evolved, to learn important principles of complexification and open endedness that may not arise in simpler environments. In Biomaker CA, the user can decide on what kind of environment and with what rules of physics the agents will live. This provides flexibility for exploring different dynamics and gradients of difficulty for survival of agents and biomes. Agents themselves can be designed freely with whatever architecture the user sees fit as well as custom mutator operators for when reproduction occurs. This allows for researching what are the most interesting and effective architectures paired with related mutators for accomplishing complexification and open endedness.

Despite the many variables among environments, some things are kept fundamental to allow for and require complexity of our lifeforms. First of all, the environments of Biomaker CA can all be classified as parts of \textit{falling-sand games} (sometimes called \textit{powder games}). Falling-sand games are characterized by grid-like worlds where the rules of the environments are mostly executed by CA rules. Every "pixel" in these worlds typically maps to a specific kind of CA with its own sets of rules. For instance, sand falls from air and forms heaps. To name a few examples, we suggest taking a look at Sandspiel \citep{bittkersandspiel} for a traditional version of such environments, and Noita for a complex videogame where this approach is used under the hoods \citep{noita2019}. Outside of videogames, researchers have also used falling-sand games for research in reinforcement learning \citep{frans2022powderworld}. In Biomaker CA, we focus on \textit{plant-like lifeforms} to strike a balance between complexity of organisms and simplicity of progress. Every fertile environment in Biomaker CA will be composed of at least five different materials (as shown in Figure~\ref{fig:basic_env}): 1) Earth, where earth nutrients can be harvested by agent roots. 2) Air, where air nutrients can be harvested by agent leaves. 3) Agent, the materials identifying living cells, each with its own logic and internal states; as we will see later, there are in truth more than one material referring to Agent. 4) Immovable, the source of earth nutrients and structural integrity. 5) Sun, the source of air nutrients. Note how there have to be \textit{both} nutrient sources in any environment for life to occur. in fact, Biomaker CA is designed to require our living organisms to constantly harvest both earth and air nutrient, each from their respective specialization of root and leaf, and pass around nutrients throughout the body so that all cells have sufficient energy to survive. A cell requires both kinds of nutrients to survive at each step and to perform more complex operations such as specialization, spawning of a new cell, and reproduction.

This condition already requires Biomaker CA's agents to be quite nontrivial and, as we will discuss, it is very hard to even have a model that survives poorly in \textit{any} environment. Moreover, in most environments organisms age and eventually die. In order for life to be preserved for longer timeframes, reproduction is necessary. Under certain conditions explained in later sections, agents that are specialized as flowers can trigger a reproduce operation that \textit{may or may not} result in a new seed being placed in the neighboring area. This seed would have a different set of parameters as randomly determined by the mutator algorithm that its parent performed over their own parameters. Hence, life is constantly changing in behavior in Biomaker CA.

Agents interact constantly with their environment, changing it at every step. Not only they harvest scarce nutrients, depleting resources for them and competing nearby plants, but also due to their effects on material types: when an agent grows roots underground, they effectively destroy Earth cells. Likewise when they grow leaves on the air, they destroy Air cells. However, while Air naturally expands to nearby empty cells, Earth cells are finite in number and can only be generated by dying agent cells that have depleted \textit{air} nutrients but still have earth nutrients inside. This causes the distribution of materials to constantly change over time and we often observe levels of earth cells to fluctuate during long simulations.

We implemented Biomaker CA to run entirely on GPUs or TPUs and we wrote it in JAX. This should allow for fast experimentation cycles as well as large scale experiments and extensions of this project. Biomaker CA is entirely open source (available at \href{https://tinyurl.com/2x8yu34s}{https://tinyurl.com/2x8yu34s}) and we welcome collaborations. in the code base we also provide example Google Colabs to perform the experiments in this article.

While an implicit goal of this project is to create complex life and understand how to build even more complex life in the future, we want to let the researchers be free with how they want to use Biomaker CA and how they want to accomplish their goals. For instance, we might be focusing on creating plants that also reproduce with variation, but other researchers might only be interested in the developmental part and not care about the reproduction part; while in this paper we focus on evolutionary strategies, researchers may want to explore the possibility of differentiable optimization for parts of the system; while we will focus on accomplishing certain results \textit{with any means necessary}, others may want to explore more open-ended, minimal viable initialization approaches to achieve open endedness; while we are not exploring the possibilities of more complex environments with more materials available, others may want to explore what kinds of materials and cell operations allow for the most interesting and complex dynamics. In essence, we want this project about open endedness to be as open ended as possible, while still making valuable design decisions that restrict the realm of possibilities in a fair way.

In order to show what is possible with Biomaker CA and to inspire researchers to collaborate with us, we will first describe the system design, then proceed with showing a variety of example configurations and kinds of experiments that can be performed. We will show how to initialize models to be able to survive in most environments, show how different design choices of rules of physics change drastically observed behaviors. We will give examples on how to evaluate models and mutators on given environments and how we can use that to evolve plants in different ways: in-environment, just by letting mutations run their course; end-to-end meta evolution, by simulating entire environments for several steps and meta-evolving the initial seed configurations; Petri dish meta-evolution, by extracting agents from the environment, placing them in to 'fake' environments and evolving them more efficiently, to later replace them into the original environments and observe significant improvements; interactive evolution, by manually selecting offspring to steer evolution in the chosen direction, allowing to get a real feel of what a mutator does to our models, and observing how these agents ultimately perform in real environments.

\section{System Design}

In this section, we will discuss the system design of Biomaker at a relative high level. This is because the project is experimental and we expect the code base to change over time, and because our open sourced code is well documented. We therefore refer to the code base for any clarification.

\begin{figure*}
  \includegraphics[width=\textwidth]{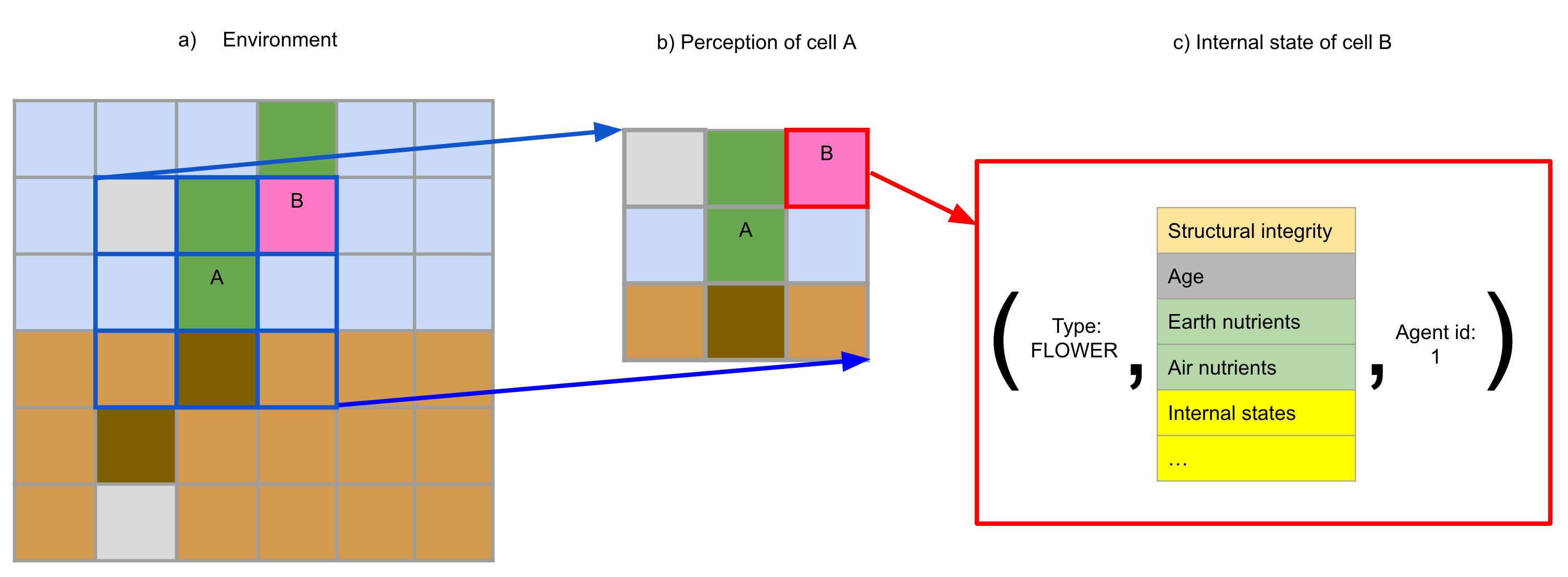}
  \caption{Schematic explaining environments and cell perceptions. a) An environment is a grid of different cells. b) Each cell perceives their 3x3 neighborhood. c) Each cell is a tuple of three elements: type (unsigned int), state (float vector), agent id (unsigned int).}
  \label{fig:env_perception_state}
\end{figure*}

\subsection{Environments, materials and configurations}

In Biomaker CA, the space where everything happens is called \texttt{Environment} (Figure~\ref{fig:env_perception_state}, sections a and c). \texttt{Environment} consists of a tuple of three grids: \texttt{type\_grid}, \texttt{state\_grid} and \texttt{agent\_id\_grid}. Each grid has the same height and width and each cell represents the space occupied by one material cell. \texttt{type\_grid} identifies the material (usually called \textit{env type}) of each position; \texttt{state\_grid} contains a vector for each cell with state information such as amounts of nutrients, agents age, structural integrity and internal states; \texttt{agent\_id\_grid} is only used for agent cells and it uniquely identifies the organism that the cell is part of. This also functions as the key for the map of all programs used in the environment, such that each organism uses their own unique set of parameters.

There are several env types available. Here we will list all the types present at the release of Biomaker CA:
\begin{enumerate}
    \item Void. The default value. Represents emptiness and it can occur when cells get destroyed.
    \item Air. Intangible material that propagates air nutrients. It spreads through nearby Void.
    \item Earth. Physical material that propagates earth nutrients. It is subject to gravity and further slides into forming heaps.
    \item Immovable. Physical material unaffected by gravity. It generates earth nutrients and structural integrity.
    \item Sun. Intangible material that generates air nutrients.
    \item Out of bounds. This material is only observed during perception, if at the edge of the world (the world is not a torus).
\end{enumerate}
Then, agent specific types start:
\begin{enumerate}
    \setcounter{enumi}{6}
    \item Agent unspecialized. The initial type an agent has when it is generated. It tends to consume little energy.
    \item Agent root. Capable of absorbing earth nutrients from nearby Earth.
    \item Agent leaf. Capable of absorbing air nutrients from nearby Air.
    \item Agent flower. Capable of performing a reproduce operation. It tends to consume more energy.
\end{enumerate}

\begin{figure}[h]
  \includegraphics[width=\columnwidth]{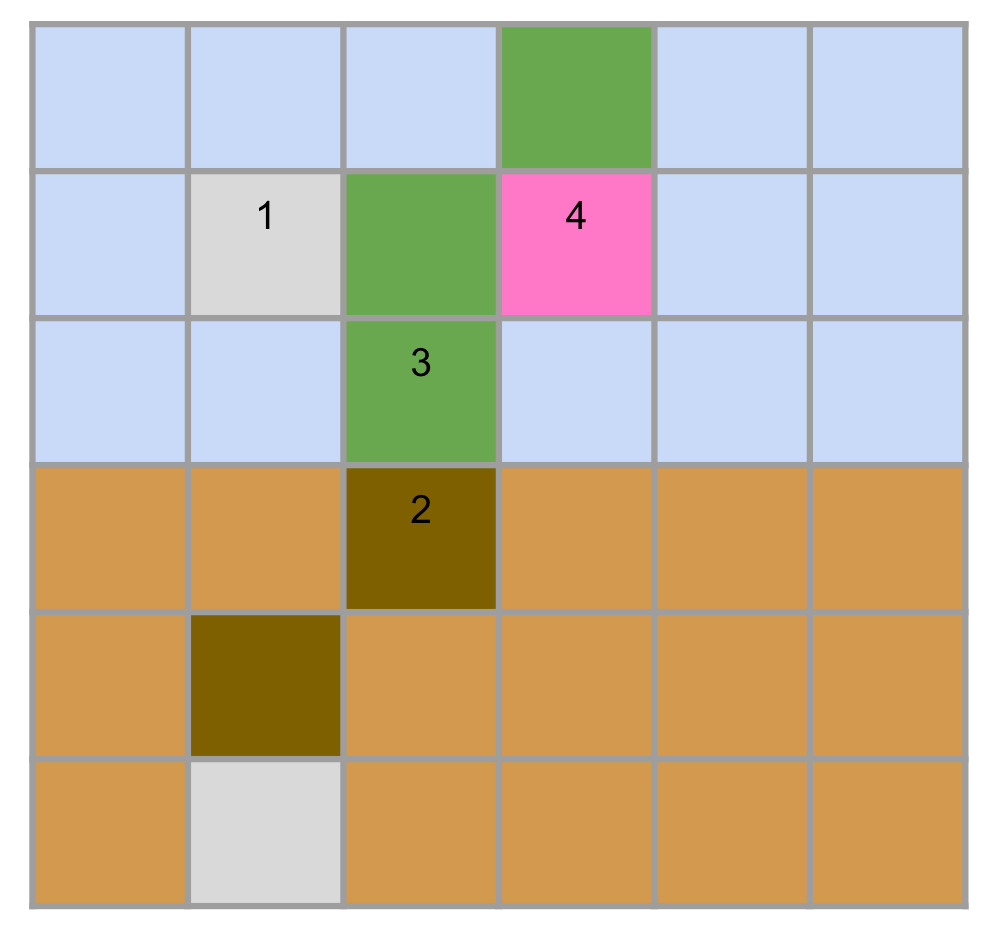}
  \caption{Agent cells. 1) Unspecialized; the initial specialization for a seed and when an agent is spawned by other cells. 2) Root; useful only next to earth cells. 3) Leaf; useful only next to air cells. 4) Flower; it can reproduce only next to air cells.}
  \label{fig:agent_description}
\end{figure}

Figure~\ref{fig:agent_description} describes the agent cells.

We will explain how these characteristics are implemented in the system whenever talking about a specific kind of logic.

Every \texttt{Environment} has to be paired with an \texttt{EnvConfig}. This represents the laws of physics of the given environment. The list of parameters is vast and we refer to the code base for an extensive list. Here, we highlight the typical topics that are tweakable with such configs: state size, absorption amounts of nutrients, maximum amounts of nutrients, costs of performing any kind of operations for agents and its details, dissipation amounts for agents (how many nutrients agents passively require for maintenance), maximum lifetime of agents.

\subsubsection{Seeds}
A seed is the initial configuration of an agent organism, as well as the configuration that gets generated whenever a reproduce operation is successful. It consists of \textit{two} contiguous unspecialized agent cells, one placed next to Air cells, one next to Earth cells (Figure~\ref{fig:basic_env}, grey cells). The reason is quite simple: in order to survive, any organism \textit{must} harvest both air and earth nutrients, hence it requires at least two different cells (excluding for degenerate cases) that have access to both kinds of nutrients. Without any seeds in an environment, no life would sprout.

\subsection{Environment logic}
These are some behaviors that happen at every step, throughout the entire grid.

\begin{figure}[h]
  \includegraphics[width=\columnwidth]{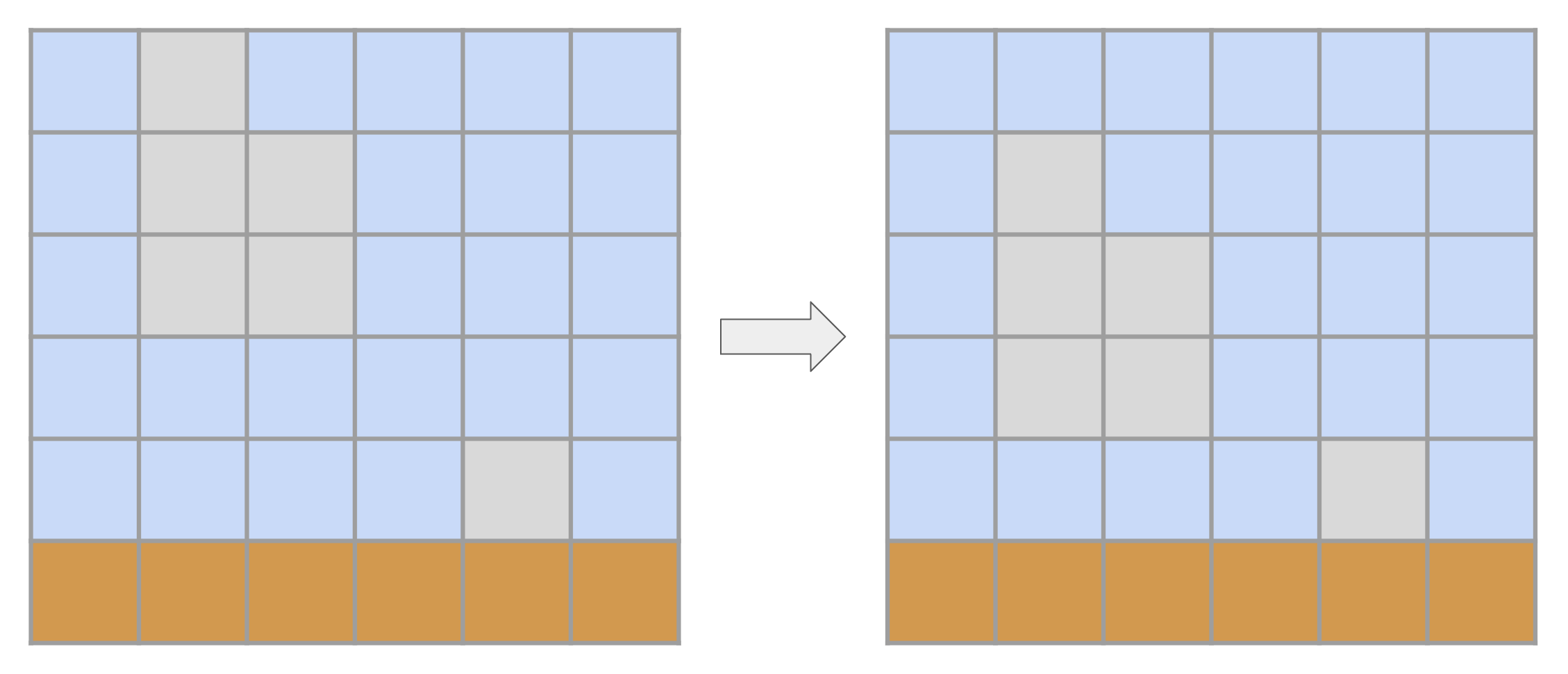}
  \caption{All cells subject to gravity fall at the same time. In the figure, agents cells fall as a unified block.}
  \label{fig:env_gravity}
\end{figure}

\textbf{Gravity.} Some materials are subject to gravity and tend to fall down whenever there is an intangible cell below them (with some caveats, as Structural integrity will describe). These materials are Earth and all Agent types. Gravity is implemented sequentially line-by-line, from bottom to up. This means that a block of detached agents/earth will fall simultaneously (Figure~\ref{fig:env_gravity}).

\begin{figure}[h]
  \includegraphics[width=\columnwidth]{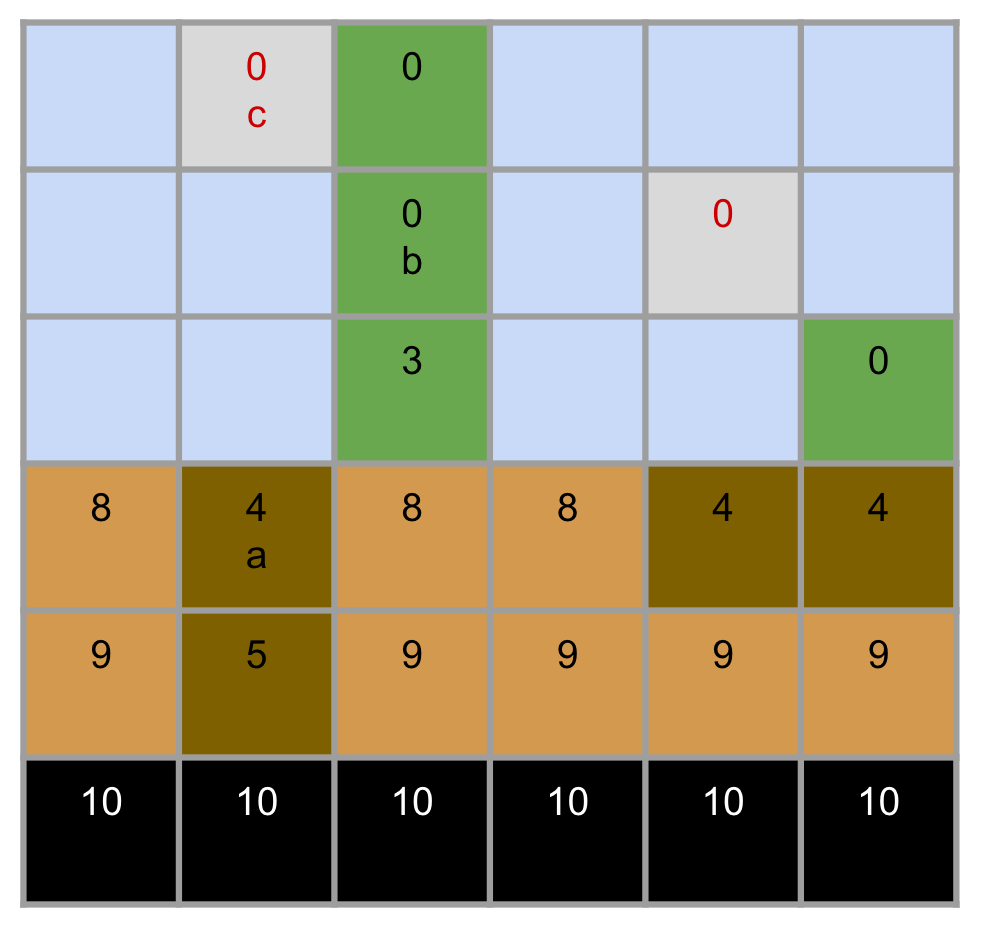}
  \caption{Example of how structural integrity would work on a small environment, where earth decays by 1, all agents decay by 5 and immutable generates 10. All cells compute their integrity by getting the maximum value of their neighbors and subtracting their decay. a) The root cell perceives an earth cell with integrity of 9, therefore its value becomes 4. b) the maximum value for the leaf's neighbors is 3, therefore its structural integrity is exhausted to zero. Now, it can fall to gravity, but since under it there is a material cell, it doesn't fall. c) this cell is also subject to gravity and there is an intangible cell below it, so it will fall to gravity.}
  \label{fig:structural_expl}
\end{figure}

\textbf{Structural integrity.} The problem with a simple handling of gravity is that plants couldn't branch out, since they would then fall to gravity. To avoid that, we devised the concept of structural integrity. In its simplest terms, this means that if an agent cell has a nonzero value of structural integrity, it will not be subject to gravity. The way we accomplish this is as follows: Immovable materials generate constantly a very high amount of structural integrity. At every step, each material cell (earth and agents) inherits the highest structural integrity in its neighborhood, decayed by a material-specific amount. For instance, earth may have a decay of 1 and agents of 5, so if the highest value of structural integrity in the neighborhood were 100, an agent would inherit a value of 95. This percolates across all earth and agents and makes it so most tree branches have a positive structural integrity and don't fall. Figure~\ref{fig:structural_expl} shows an example of how structural integrity propagates. Note that earth only propagates structural integrity but is still always subject to gravity, and that if a plant were to be severed during its lifetime, it would take some time for structural integrity to set the cut value's structural integrity to zero. Therefore, we generally perform multiple iterations of structural integrity processing per environmental step.

\textbf{Aging.} At every step, all cells age. This increases a counter in their state values. As the next section will discuss, agents that have lived past their half maximum lifetime will start to dissipate linearly increasing extra energy per step. The only way to reset the aging counter is to reproduce: new seeds (with new agent ids) will have their age counter set to zero.

\textbf{Energy processing.} Immutable and Sun materials constantly generate new nutrients that are diffused to nearby earth and air cells respectively. These cells, in turn, diffuse these nutrients among themselves. This creates an issue of percolation: if an earth/air cell is unreachable by other earth/air cells with nutrients, they will not receive any nutrients. At the same time that nutrients are diffused, roots and leaves neighboring with earth/air cells will harvest a fixed amount of nutrients (if available) from them. Afterwards, all agents dissipate a fixed amount of nutrients. Moreover, if they have reached past their half lifetime, they lose extra nutrients with an ever increasing amount over time. If they don't have enough nutrients, they die. In case of death, if earth nutrients are left, the agent turns into earth. If air nutrients are left, the agent turns into air. Otherwise, the agent becomes void.

\subsection{Perception}
No matter the kind of cell, be it a material or an agent, they can only perceive a limited amount of data. This is their 3x3 neighborhood in the \texttt{Environment} (Figure~\ref{fig:env_perception_state}.b). Therefore, each cell perceives their respective 3x3 neighborhood of \texttt{type\_grid}, \texttt{state\_grid} and \texttt{agent\_id\_grid}. Note that this implies that agent cells can perceive the unique identifiers of agent id to distinguish themselves from different organisms. This feature is entirely optional, and the user can choose to limit agents perceptions to only type and states of neighbors. All the experiments we show in this paper are however using models that do distinguish between different organisms, because we observed that otherwise the systems would generally converge to have several multi-agent-id organisms, and we wanted to explore pure organisms.

\subsection{Cell operations}
In this section we discuss how cell type logic is implemented. This includes how to define the behavior of materials as well as what is possible for agents to do, and give a peek at how it is implemented at the low level.

There are three kinds of operations that can be defined in this framework: \texttt{ParallelOp}, \texttt{ExclusiveOp} and \texttt{ReproduceOp}.

\subsubsection{Parallel operations}

\begin{figure}[h]
  \includegraphics[width=\columnwidth]{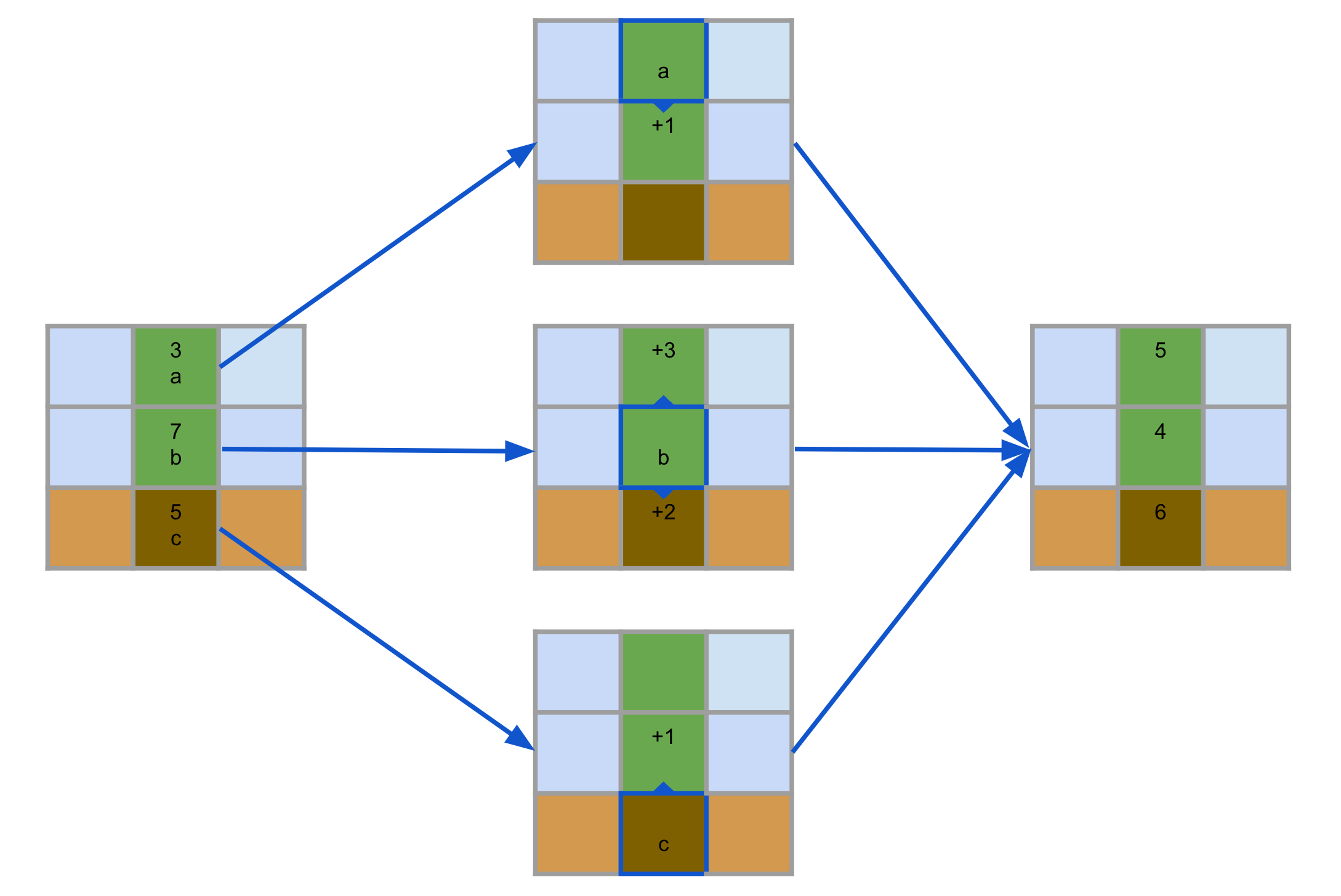}
  \caption{Example how exchange of nutrients occurs on a parallel operation. Each cell decides how much to give away to each neighboring agent. This gets accumulated. This process is repeated twice, once for each nutrient type (earth and air).}
  \label{fig:synchronous_example}
\end{figure}

Parallel operations (\texttt{ParallelOp}s) are kinds of operations that are safe to be performed all at the same time, without having any issues of conflict. In practice, we use them only for agent behaviors. Agents can perform 3 kinds of parallel actions: updating their own internal states, changing their own specialization, and distributing nutrients to neighboring agents. Since these operations neither create nor destroy anybody else, they are safe to be performed in parallel.

\texttt{ParallelOp}s – and as we will see \textit{all} \texttt{Op}s – are meant to be processed and aggregated by the system. Therefore, they need to be valid operations that don't break the laws of physics. Alas, we cannot trust our agent logics to not exploit everything that they can for survival, therefore we create an interface that agents use, \texttt{ParallelInterface}, as the output of their functions. These interfaces are then sanitized and converted to always valid \texttt{ParallelOp}s.

These agent operations, in particular, have these restrictions: 1) Energy cannot be created out of thin air, and an agent can only give nutrients away (never ask for nutrients), as it can be seen in the example in Figure~\ref{fig:synchronous_example}; 2) Specialising has a cost in nutrients, and if the cell doesn't have enough nutrients, specialization doesn't work. Finally, there is no restriction on what an agent does with its own internal states.

A technical note: We could theoretically use \texttt{ParallelOp}s to perform energy operations of all cells, including diffusion, energy dissipation and death. We don't do that for efficiency, but future refactoring may be performed if deemed useful.

\subsubsection{Exclusive operations}

\begin{figure}[h]
  \includegraphics[width=\columnwidth]{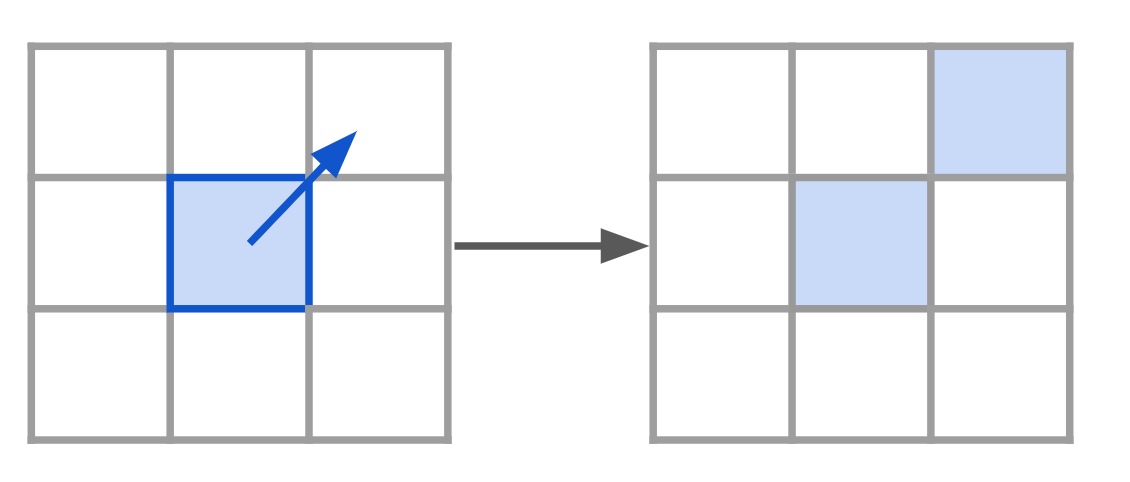}
  \caption{Air exclusive operation: choose one random Void cell nearby and create a new Air cell there. Note that every ExclusiveOp can only act on maximum one neighboring cell.}
  \label{fig:air_conc_op}
\end{figure}

\begin{figure}[h]
  \includegraphics[width=\columnwidth]{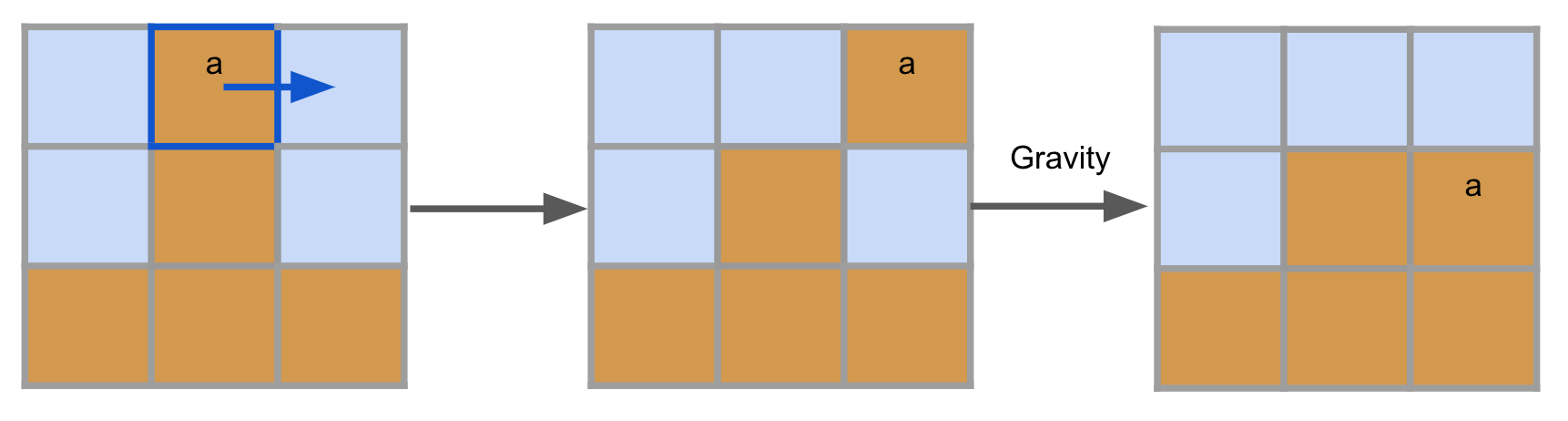}
  \caption{Earth exclusive operation: if you are stacked on top of other cells and you could fall sideways, move to the side. Then, gravity will independently make the cell fall.}
  \label{fig:earth_conc_op}
\end{figure}

\begin{figure}[h]
  \includegraphics[width=\columnwidth]{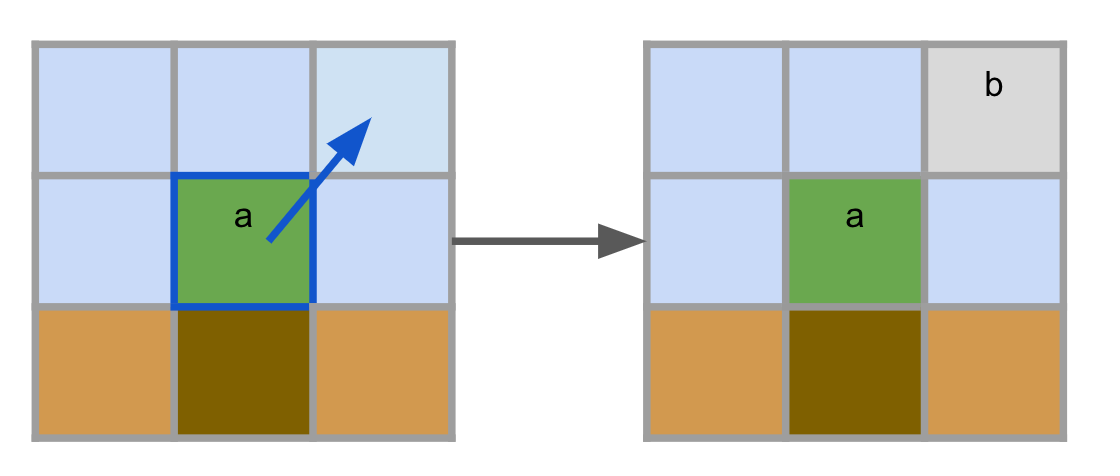}
  \caption{Spawn operation: agent a selects an empty cell to create agent b. Both agents share the same agent ids and the nutrients in agent a are split in two and some are given to b. Agent b is always Unspecialized.}
  \label{fig:spawn_op}
\end{figure}

Exclusive operations (\texttt{ExclusiveOp}s) are operations that are potentially not safe to be performed all at the same time. This typically implies the change of cell type acting on a neighboring cell. For instance, if an Earth cell wants to slide into position A filled with Void, and Air cell wants to duplicate itself in position A, and an Agent wants to spawn a new cell in position A, only one of these operations can be performed at any given time. Moreover, the effect of whatever operation is performed would change the feasibility of other operations. To solve these conflicts, we write our exclusive operations with a paradigm of \textit{atomic commits}: 1) each cell makes a proposal of what it would like to happen to (always up to) one neighbor. With that it also states what would happen to itself should this operation happen. For instance, moving Earth cells would state to create an Earth cell in position A and replace its own position with the content of position A, while Agent cells performing a spawn operations would state how many nutrients to subtract from itself to give to the new cell and how to modify its own internal states. 2) We accumulate all \texttt{ExclusiveOp}s that are requested to be performed for each target cell. 3a) If only one cell is targeting this cell, we execute that operation, updating both the target and actor cell. 3b) if more than one cell wants to act on the target, we randomly break the conflict executing only one of these operations.

At release time, we have only 3 kinds of exclusive operations: Air spreads through Void (Figure~\ref{fig:air_conc_op}); Earth performs a exclusive version of the typical falling-sand algorithm, shifting to the side if it can fall sideways (Figure~\ref{fig:earth_conc_op}); Agents can choose to perform a "Spawn" operation, generating a new cell in the neighborhood with the same agent's id (Figure~\ref{fig:spawn_op}).

While we can trust our own code for the operations of Earth and Air cells and write it to output \texttt{ExclusiveOp}s directly, we cannot trust agents. So, here as well we use a \texttt{ExclusiveInterface} to validate a cell's wish to perform Spawn and convert it to a \texttt{ExclusiveOp} in case. Note that, currently, agents can only perform Spawn as a exclusive operation. If agents were to be capable of performing more than one operation, the \texttt{ExclusiveInterface} would have to be responsible for making a decision on what to perform and convert that choice in a \texttt{ExclusiveOp}. For instance, if we wanted to allow agents to "move", we would have to add that option into the ExclusiveInterface. Specifically for "movement", we discourage its implementation at least while focusing on plant-like organisms. The reason is that CA that want to preserve a connected organism are not very friendly with this kind of movement operation: if a cell moves, the rest of the organism stays put.

For the Spawn operation, there are certain restrictions: 1) agents cannot spawn a cell everywhere. Currently, only Void, Air and Earth cells are valid recipients of Spawn. 2) Spawn costs nutrients, and if the agent doesn't have enough energy, the operation fails.

\subsubsection{Reproduce operations}

\begin{figure}[h]
  \includegraphics[width=\columnwidth]{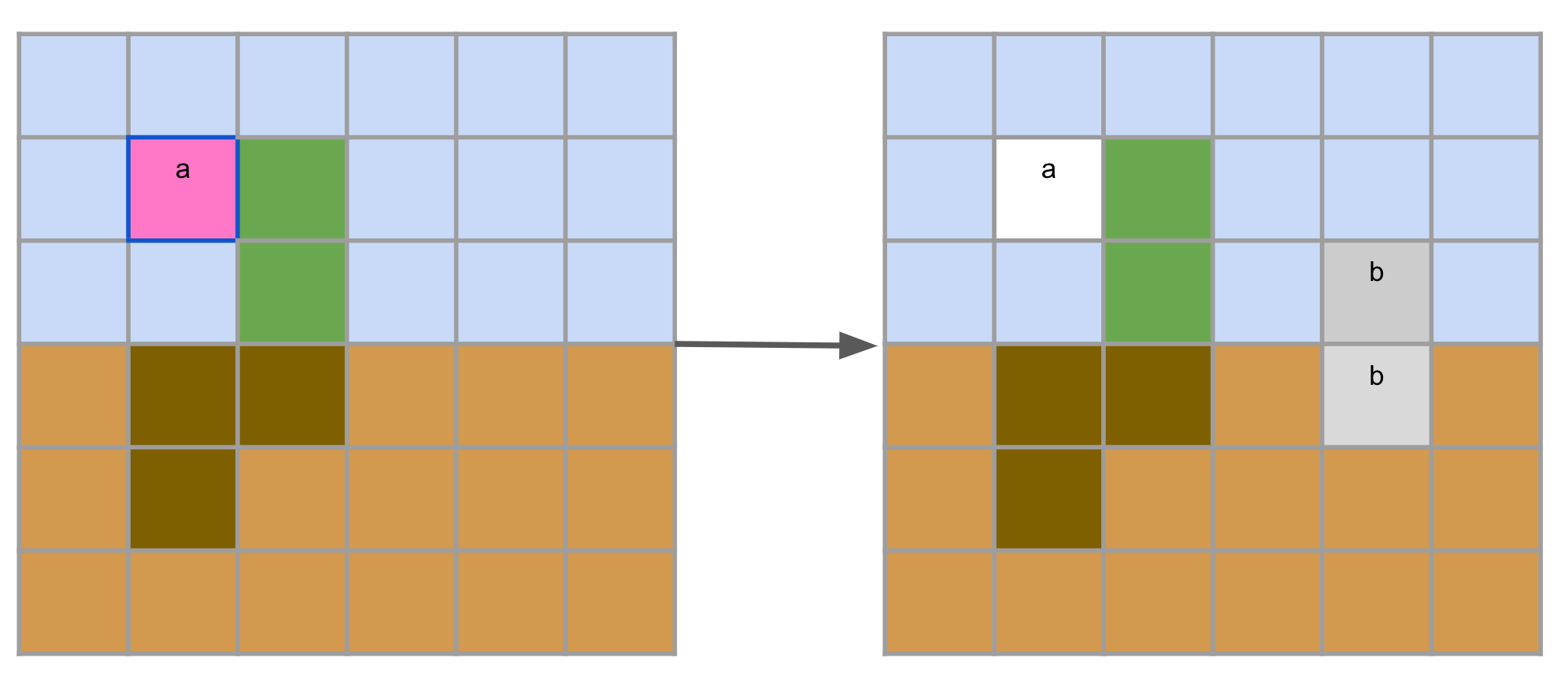}
  \caption{Reproduce operation: agent a triggers reproduction and gets destroyed and replaced with a Void cell. If the reproduction succeeds, a new seed (the two cells b) will be generated in a random neighborhood, containing the remainder of agent a's nutrients after the cost of the operation is subtracted.}
  \label{fig:reproduce_op}
\end{figure}

Organisms have to eventually reproduce. They do so by specialising into flowers and trying to perform a \texttt{ReproduceOp}. As usual, we created an interface called \texttt{ReproduceInterface} that agent functions use and we take care of validating this operation and converting it to a \texttt{ReproduceOp}.

Reproduce operations are quite different from other kinds of operations. First of all, the requirement for creating a valid \texttt{ReproduceOp} is quite simple: just have enough nutrients for the cost of the operation and be a flower. Then, however, all the remaining nutrients will be used for generating a new seed. So, if the flower didn't store enough nutrients, it is very likely to generate an infertile seed.

Generating a valid \texttt{ReproduceOp} does not however ensure that a new seed will be generated. For it to happen, first the flower needs to be adjacent to Air cells (the more, the more likely it is to succeed), so underground flowers will never generate seeds. Then, for each step, only a very small number of \texttt{ReproduceOp}s get selected (the default value is 2, but it can be changed from \texttt{EnvConfig}s), therefore many flowers will have to wait a long time for them to be picked up. This, while it was implemented for efficiency concerns, has the nice parallel of having to compete for limited environmental resources aiding reproduction (for instance, virtual bees or birds).

If a flower is picked up for reproduction, it gets immediately deleted and a Void cell replaces it. Then, the system tries to place a new seed with a uniquely new agent id in the neighborhood (Figure~\ref{fig:reproduce_op}). However, this also can fail. It fails if either of the two is true: 1) there is no \textit{fertile ground} nearby (a free space where there are contiguous Air and Earth cells stacked); 2) we have already reached the limit of available number of unique agent ids for the environment.

The latter requirement is dependent on the fact that whenever an organism reproduces, it generates \textit{new parameters} that are uniquely identified and stored in a structure keyed by the agent id. So, if the maximum number of programs is 64, we cannot spawn a new agent if there are already 64 unique organisms in the system. This limitation is entirely of spatial complexity: our implementation does not get slowed down by having however many possible potential programs, but the RAM may not handle it.

This restriction on the number of programs is also assuming that we are performing reproduction with variation. This is the default behavior, which requires a Mutator as we will see in a later section. However, this too can be disabled. In this case, reproduction happens without variation and there is no limit to the number of unique agent ids that an environment can handle.

\subsection{Agent logic}

Probably the most important design choice that a researcher has to make in Biomaker CA is the choice of the agent logic and, if interested in reproduction with variation, its respective mutation strategy (mutator).

The agent logic is a set of three different functions and its respective parameters (in short, models) that define the behavior of a given agent with respect to its parallel, exclusive and reproduce operations. All three operations accept the same kinds of inputs: the cell's perception (neighboring types, states and agent ids) and a random number (in case the model wants to act stochastically). The outputs are respectively a \texttt{ParallelInterface}, \texttt{ExclusiveInterface} and \texttt{ReproduceInterface}.

This approach leaves the researcher free to implement whatever kind of logic they want in order to create their agent operations. The only requirement is for it to be vectorizable in JAX, as we will vectorize these operations for each cell in the environment grid. This design choice implies that all agents in a given environment will use the same agent logic, and their diversification can only occur through changes in their parameters. This can be seen as being yet another set of rules of physics that agents have access to. In practice, this approach makes it trivial to vectorize our models and it scales for large environments. However, this approach is not a strict requirement, and it is possible to extend this framework to have different agents using different agent logics, at the cost of multiplying the computation cost by the number of unique agent logics.

It is crucial to understand how unlikely it is for a random agent logic with random initial parameters to actually generate a fertile organism, capable of growing and reproducing. In fact, Biomaker CA is so complex that most trivial models or initialization will undoubtedly fail. This is because in order to reproduce starting from a seed of two unspecialized cells, the organism needs to at minimum be capable of: 1) specialize accordingly into root and leaf cells; 2) distribute nutrients to neighbors smartly enough to make sure that no cell (or few unimportant ones) dies; 3) perform spawn operations to grow in size and accumulate more nutrients and more importantly create a flower; 4) store extra nutrients in the flower and trigger reproduction. Most initializations and architecture will fail from the start and won't be able to build up anything even through meta-evolution because they would constantly receive zero signals of improvements.

We therefore provide the researchers with two initial agent logics, one with ~300 parameters and one of more than 10000, that are hand-designed to have an initialization that can create fertile plants in most environments. Bootstrapping an ALife environment with pretrained agents is not something novel: especially when working with Minimal Criterion environments such as these, researchers seem to need to bootstrap agents in some ways \citep{soros2014,brant2017as}. We chose to hand-write a parameter skeleton to accomplish this goal. This approach is consistent with our perspective that we are more than willing to interact with the environment at specific stages of development if this results in a tremendous speed up in computation or a better resulting complexification. In this paper, we will call "minimal" the small architecture and "extended" the larger, since the latter is indeed an extension of the minimal one (it strictly contains more operations and parameters). The minimal architecture does not ever modify an agent's internal states, while the extended architecture can modify them at will and act based on internal states as well.

We will later show results by using these basic architectures, but we stress that these architectures are clearly just a starting point and there is much more to explore for figuring out what are the most evolvable architectures and mutators for open endedness and growing complexity.

\subsection{Mutators}

The initial parameters of the agent logic are just the starting point. One can mutate and evolve them in several different ways, but if they want to allow for in-environment mutation through reproduce operations, they must use \textit{mutators}. A mutator is a very simple function that takes as input some parameters and a random number, and with that they generate a new set of parameters. This is the function that gets executed whenever a reproduce operation is triggered.

We can roughly distinguish between two kinds of mutators: stateless and stateful. A stateless mutator does not have any parameters by itself. In this paper, we refer to "basic" as a stateless mutator that mutates parameters by sampling the new ones from a gaussian with a fixed standard deviation, and has a 20\% chance of updating each parameter. A stateful mutator adds parameter to the agent logic. We instead refer to "adaptive" when using a stateful mutator that doubles the amounts of parameters of the agent logic so that each and every parameter has its own standard deviation for mutation. Each parameter also has a 20\% chance of update but crucially also their standard deviation gets in turn randomly updated with a certain chance and standard deviation. We refer to the code base for more details.

When a mutator generates new parameters, it generates new parameters both for the agent logic and the mutator's. This shows how closely related mutators are to the agent logic. Ultimately, the quest for open endedness and complexification requires to figure out the interplay between modifying a genotype through mutation strategies to result in new genotypes that express interesting new phenotypes.

Note that mutators don't receive extra information such as ranking of a population or fitness values. The only information that they receive is that reproduction was triggered. We believe that research within the area of what are the best mutators and genotype-to-phenotype mapping (in our case, agent logic) is at its embryonic stages, and we encourage researchers to figure out the best working pairs of mutators and agent logic for general complexification.

\section{Examples and experiments}

In this section, we will show four different examples of configurations (pairs of \texttt{Environment} and \texttt{EnvConfig}). The laws of physics defined by \texttt{EnvConfig} can be used for many different \texttt{Environment}s, creating in principle different configurations every time. However, to simplify the nomenclature, when the only change between two configurations is the size of the \texttt{Environment}, we consider them the same configuration, albeit we specify the size of the environment. For instance, Figure~\ref{fig:hero_figure} shows two different sizes for the configuration called "persistence".

\begin{figure}[h]
  \includegraphics[width=\columnwidth]{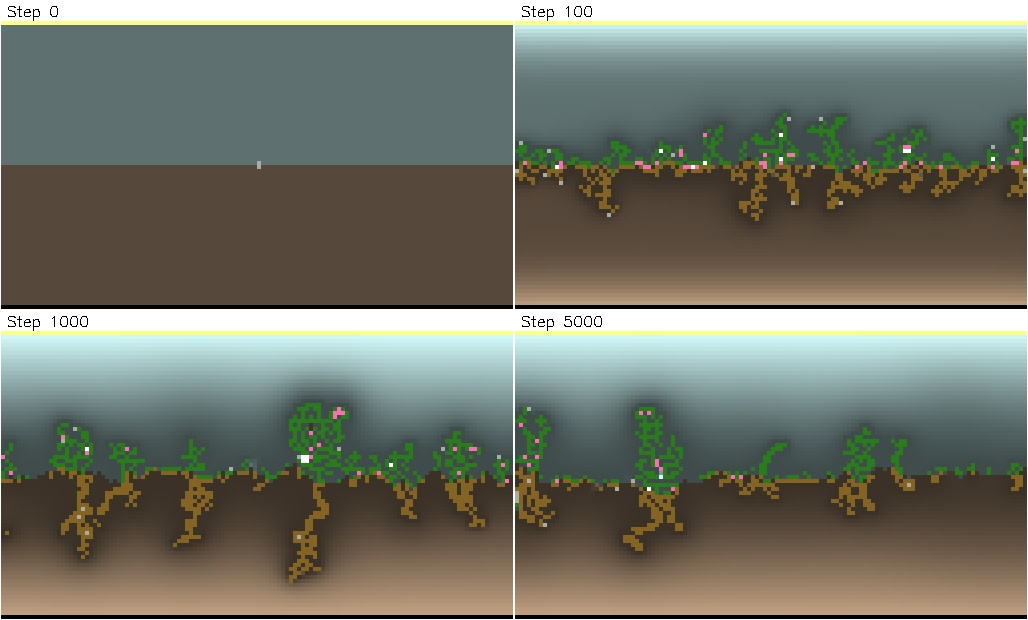}
  \caption{Example of a run of "persistence" with a minimal agent logic and a basic mutator.}
  \label{fig:persistence_run_example}
\end{figure}

Persistence is a configuration where agents age very slowly (max age is 10000). This means that agents will eventually die, but it takes so long that they have the time to grow as complex as they desire. Moreover, the cost of maintaining a cell (dissipation) is very low and the cost of specialization is low. However, the cost of spawn and reproduce are very high. In practice, this configuration is very easy and there often appear big, persisting plants. Figure~\ref{fig:persistence_run_example} shows an example run of persistence for 5000 steps. Note how step 0 is scarce of nutrients and always starts with a single seed in the center. This will be true for all configurations in this article. Also, specifying a configuration is not enough for uniquely identifying an example run. In fact, we also need to specify what agent logic and mutator we used. For Figures \ref{fig:hero_figure} and \ref{fig:persistence_run_example}, we used a minimal agent logic and a basic mutator.

\begin{figure}[h]
  \includegraphics[width=\columnwidth]{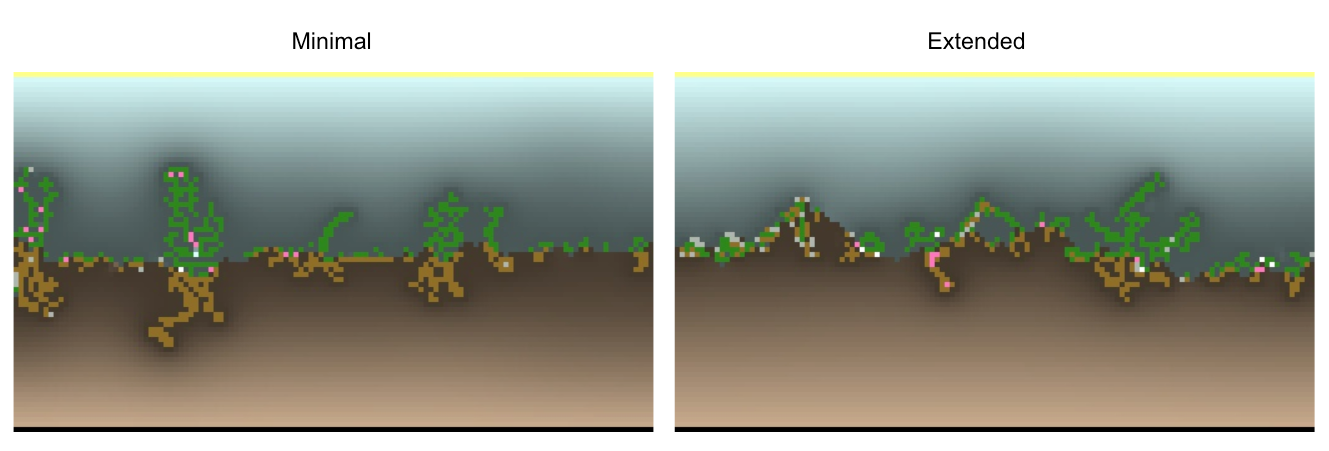}
  \caption{Comparison of the state of "persistence" after 5000 steps. Left: a minimal agent logic. Right: an extended agent logic. Both have a basic mutator, but the standard deviations are 0.01 and 0.001 respectively.}
  \label{fig:persistence_comparison}
\end{figure}

To show an example of how a different agent logic changes completely the results of the simulations, Figure~\ref{fig:persistence_comparison} shows the status of two runs, with a minimal and an extended agent logic respectively and both using a basic mutator, after 5000 steps. The extended biome is much more diverse even if we used a lower standard deviation for the mutator of the extended version.

\begin{figure}[h]
  \includegraphics[width=\columnwidth]{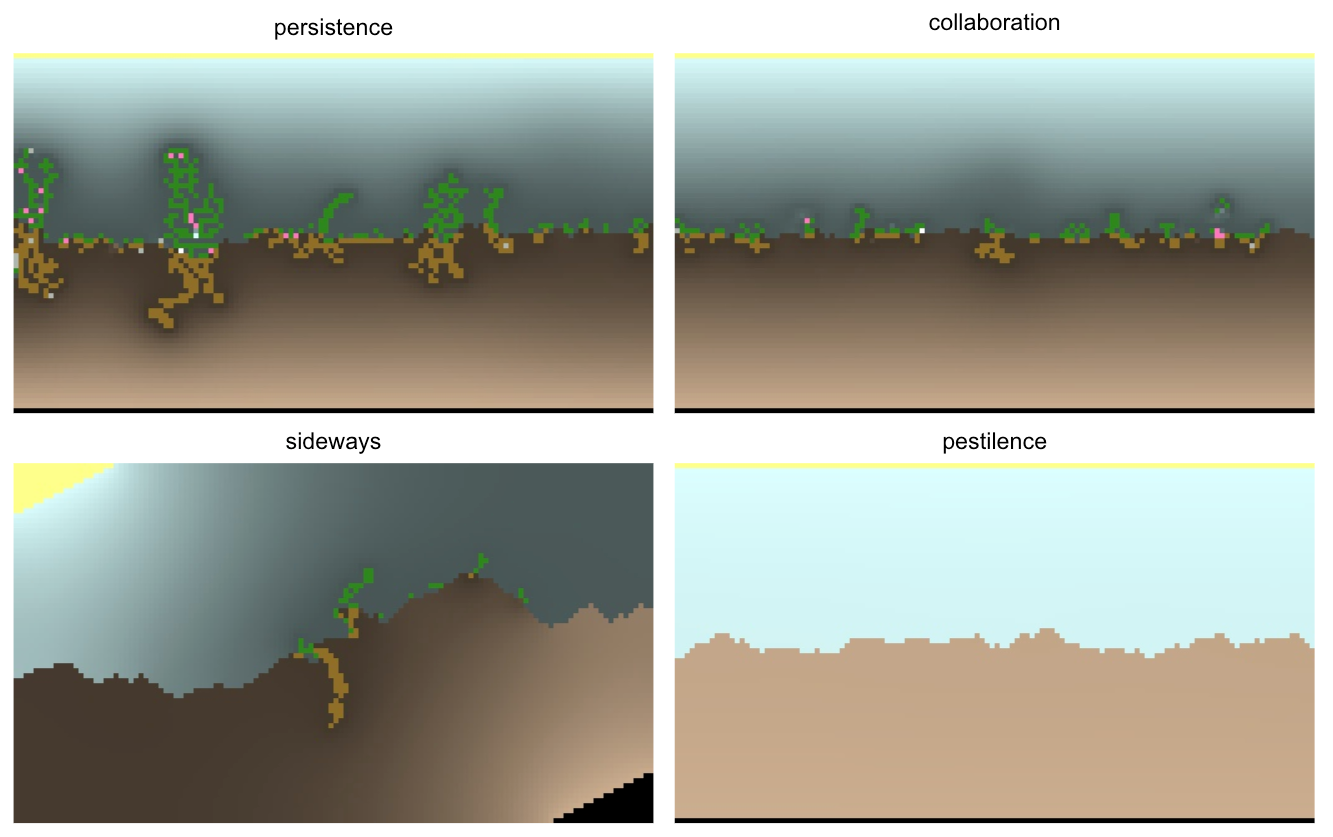}
  \caption{Comparisons of different configurations after 5000 steps. All configurations use a minimal agent logic with a basic mutator.}
  \label{fig:configurations_comparison}
\end{figure}

Now we will describe three more configurations as examples. For a thorough description of any configuration, we refer to the code base. In "collaboration" (Figure~\ref{fig:configurations_comparison}, top right), agents die after 100 \textit{million} steps. In practice, they don't age. However, this environment is harder than persistence due to its very high dissipation value and its bigger specialization cost. "sideways" instead shows how different a configuration can be even if it has the same \texttt{EnvConfig} of persistence. This is because the nutrient generators are not uniformly distributed at the top and bottom of the environment, but instead they are only present in north-west and south-east, making the environment very scarce with resources (Figure~\ref{fig:configurations_comparison}, bottom left). As we will see later, this environment is deceptively hard. Finally, in "pestilence" agents age very quickly (max age is 300) and specialization is costly (Figure~\ref{fig:configurations_comparison}, bottom right). This makes this environment very complex for our agent logics. In fact, Figure~\ref{fig:configurations_comparison} shows the result of running for 5000 steps the four configurations described, where we used the same minimal agent logic and basic mutator. Pestilence is the only one where the biome got extinct.

This brings up a few questions. First of all, how likely is it for pestilence to go extinct after a given number of steps? Likewise for the other configurations, do they ever go extinct? Is there a way to evaluate the performance of a given pair of agent logic and mutator over a configuration to understand what works best?

\begin{figure*}
$$
\begin{array}{ c | c | c | c | c }
\text{config name} & \text{logic} & \text{mutator} & \text{total agents} & \text{extinction \%} \\
\hline
 \text{persistence} & \text{minimal} & \text{basic} & 563470 \pm 46286  & 0 \\ 
 \text{persistence} & \text{minimal} & \text{adaptive} & 549518 \pm 21872 & 0 \\ 
 \text{persistence} & \text{extended} & \text{basic} & 462199 \pm 52976 & 0 \\ 
 \text{persistence} & \text{extended} & \text{adaptive} & 448378 \pm 56991 & 0 \\ 

 \text{collaboration} & \text{minimal} & \text{basic} & 147008 \pm 1607  & 0 \\ 
 \text{collaboration} & \text{minimal} & \text{adaptive} & 146768 \pm 3028 & 0 \\ 
 \text{collaboration} & \text{extended} & \text{basic} & 129668 \pm 23532 & 6.25 \\ 
 \text{collaboration} & \text{extended} & \text{adaptive} & 127100 \pm 23876 & 6.25 \\ 
 
 \text{sideways} & \text{minimal} & \text{basic} & 296927 \pm 14336  & 0 \\ 
 \text{sideways} & \text{minimal} & \text{adaptive} & 293534 \pm 14377 & 0 \\ 
 \text{sideways} & \text{extended} & \text{basic} & 259805 \pm 18817 & 0 \\ 
 \text{sideways} & \text{extended} & \text{adaptive} & 254650 \pm 32019 & 0 \\ 

 \text{pestilence} & \text{minimal} & \text{basic} & 151439 \pm 82365  & 68.75 \\ 
 \text{pestilence} & \text{minimal} & \text{adaptive} & 165650 \pm 70653 & 62.5 \\ 
 \text{pestilence} & \text{extended} & \text{basic} & 171625 \pm 65775 & 43.75 \\ 
 \text{pestilence} & \text{extended} & \text{adaptive} & 156197 \pm 59918 & 56.25
\end{array}
$$
  \caption{Evaluations of different candidate triplets of (configuration, agent logic, mutator) run for 1000 steps.}
  \label{fig:eval_configs}
\end{figure*}

\subsection{Evaluation methods}

To answer these questions, we propose some methods for evaluating a candidate triplet of (configuration, agent logic, mutator) in a way that we can create a fitness function. Having a fitness function will also allow us to perform some meta-evolution, as we will see in later sections.

We start by creating a function that runs a candidate for 1000 steps to track two metrics: how many agent cells are occupied throughout the entire run, and whether at the end of the 1000 steps the agents got extinct. The total number of agents will be our crude metric for identifying how much of the environment agents are taking over, while the extinction flag will tell us whether the candidate "failed" to survive.

Since we implemented everything in JAX, we can easily parallelize the computation of these metrics. Therefore, we will run each candidate 16 different times each and accumulate the statistics. Figure~\ref{fig:eval_configs} shows the result of these evaluations. As expected, persistence is the easiest environment, where the total number of agents is much higher than the rest and no extinction ever occurs. Second place is sideways where, while it shares the physics of persistence, it results harder for agents to take over more space in the environment. Collaboration and pestilence seem to be much harder and we can clearly see how the extinction rate on pestilence is very high – but it is not 100\%.

It also appears that for all configurations except pestilence the minimal logic is better than the extended logic. Keeping in mind that these have to be compared alongside the mutators used, and that perhaps our mutators are suboptimal for extended models, this result is quite undesirable since the extended logic is a literal superset of the minimal one and its initialization results in an identical behaviour of the minimal model. Alongside the observation that the adaptive mutator seems to be slightly underperforming the basic mutator, we believe that there is a vast amount of research that needs to be done in the area of figuring out the best agent logics and mutators.

\subsubsection{Deceitful evaluations}
Actually, this evaluation method may be insufficient for configurations where there are significant long-term environmental changes. If we only evaluate the status of a configuration for up to 1000 steps, we would be unaware of the case where agents would get extinct after that. This is the case for sideways (and sideways only, among the configurations explored). In sideways, dying agents will drastically change the amount and placement of Earth cells over time, and agents can all get extinct at some point.
\begin{figure}[h]
$$
\begin{array}{ c | c | c | c }
\text{logic} & \text{mutator} & \text{total agents} & \text{extinction \%} \\
\hline
 \text{minimal} & \text{basic} & 1M \pm 230k  & 31.25 \\ 
 \text{minimal} & \text{adaptive} & 1.1M \pm 191k & 18.75 \\ 
 \text{extended} & \text{basic} & 644k \pm 204k & 81.25 \\ 
 \text{extended} & \text{adaptive} & 500k \pm 164k & 81.25
\end{array}
$$
  \caption{Evaluations of different candidate pairs of (agent logic, mutator) run for 10000 steps on the "sideways" configuration.}
  \label{fig:eval_sideways}
\end{figure}

\begin{figure}[h]
  \includegraphics[width=\columnwidth]{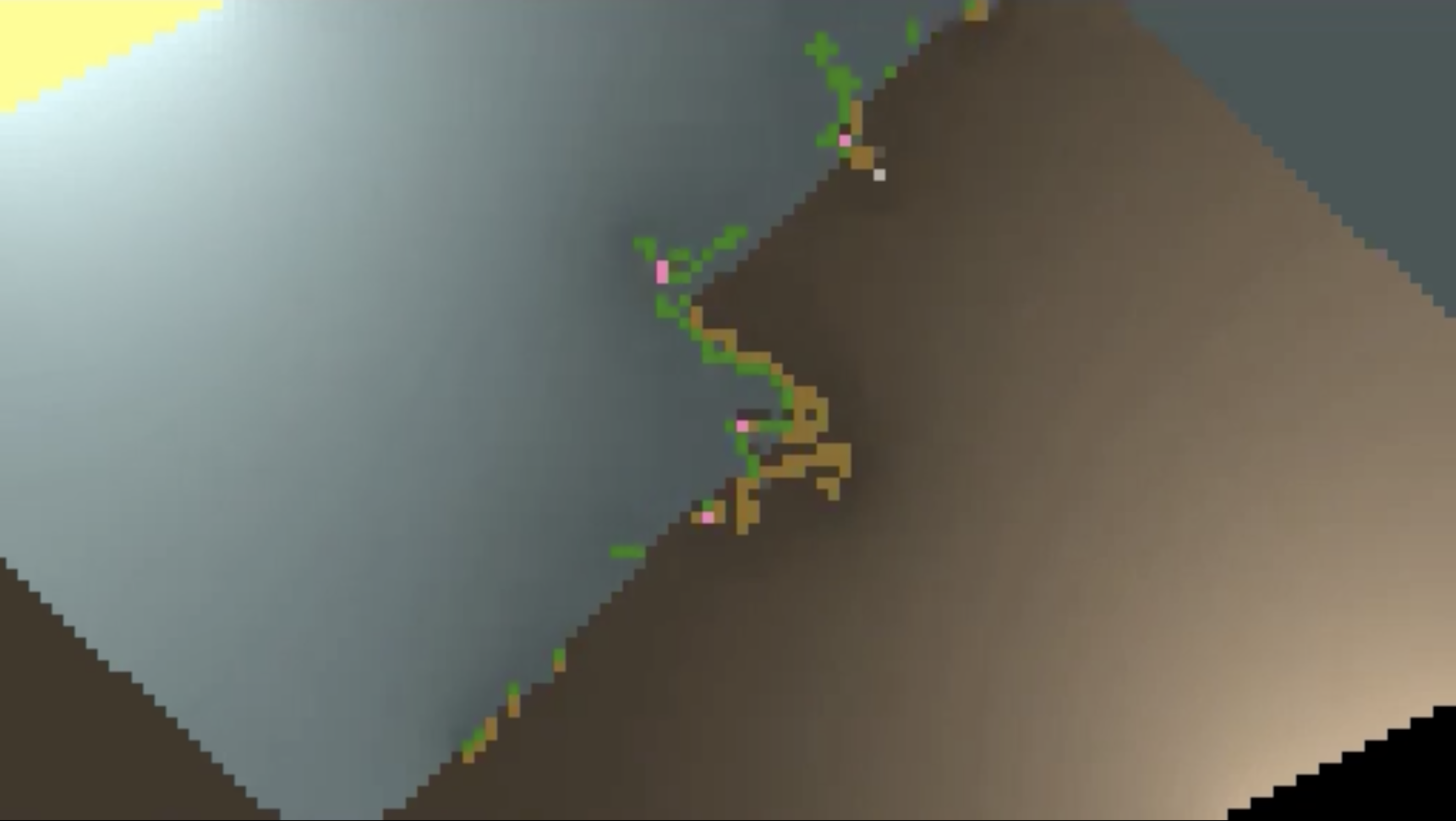}
  \caption{Snapshot of the state of a "sideways" run after roughly 38k steps with a minimal agent logic and an adaptive mutator.}
  \label{fig:sideways_late}
\end{figure}

Figure~\ref{fig:eval_sideways} demonstrates how performing our previous evaluation for 10000 steps shows that every kind of agent logic and mutation pair can get extinct, and that the minimal configuration with adaptive mutators seems significantly better than the rest. This is the first time that we observe that adaptive mutators perform better than basic ones. To give a sense of what is the difficulty of the environment, Figure~\ref{fig:sideways_late} shows a run of a minimal logic with adaptive mutators after roughly 38000 steps. The environment has become a gigantic heap of earth and survival is much harder. In that simulation, life will get extinct shortly after.

\subsubsection{Focusing on pestilence}
There are many problems that we could try to solve in the configurations that we have introduced. In this paper, we will focus on solving the very high extinction rate observed in pestilence. We will take the best performing candidate (extended logic with basic mutator) and see how far we can get with improving our chosen metrics. Throughout the rest of this article, we will evaluate our models while running them for 1000 steps. As we have seen, this is not generally a robust approach, but it will be enough for pestilence specifically. Evaluating each candidate for 10000 steps requires a significantly larger amount of resources, so we chose to avoid it when we know that it is not needed.

\begin{figure*}
  \includegraphics[width=\textwidth]{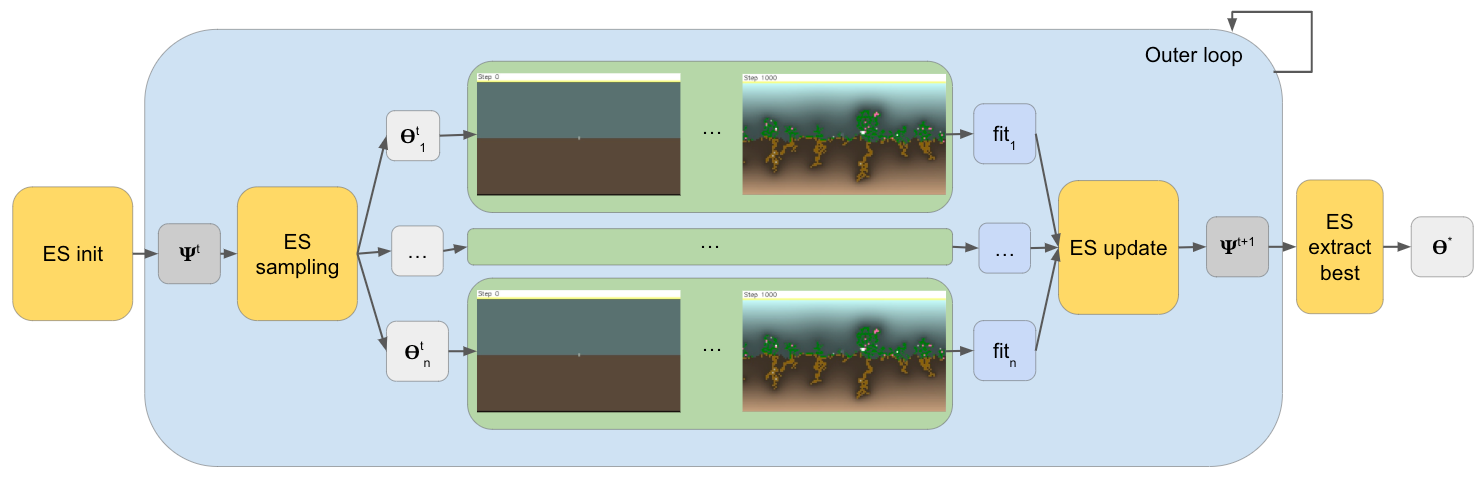}
  \caption{Diagram explaining meta-evolution. First, we initialize some parameters $\Psi^0$ containing agents initialized parameters and some ES-specific parameters. For each outer step t+1, we sample a population of agent parameters $\Theta^t_i$ using $\Psi^t$. These parameters are each used to simulate a run for several inner steps (1000 in the example). A fitness is extracted by each of these runs and the meta-evolutionary strategy aggregates this information to generate the new parameters $\Psi^{t+1}$. When the outer loop finishes, we extract the best agent parameters $\Theta^*$.}
  \label{fig:meta_evolution_diagram}
\end{figure*}

\subsection{In-environment evolution}

\begin{figure}[h]
  \includegraphics[width=\columnwidth]{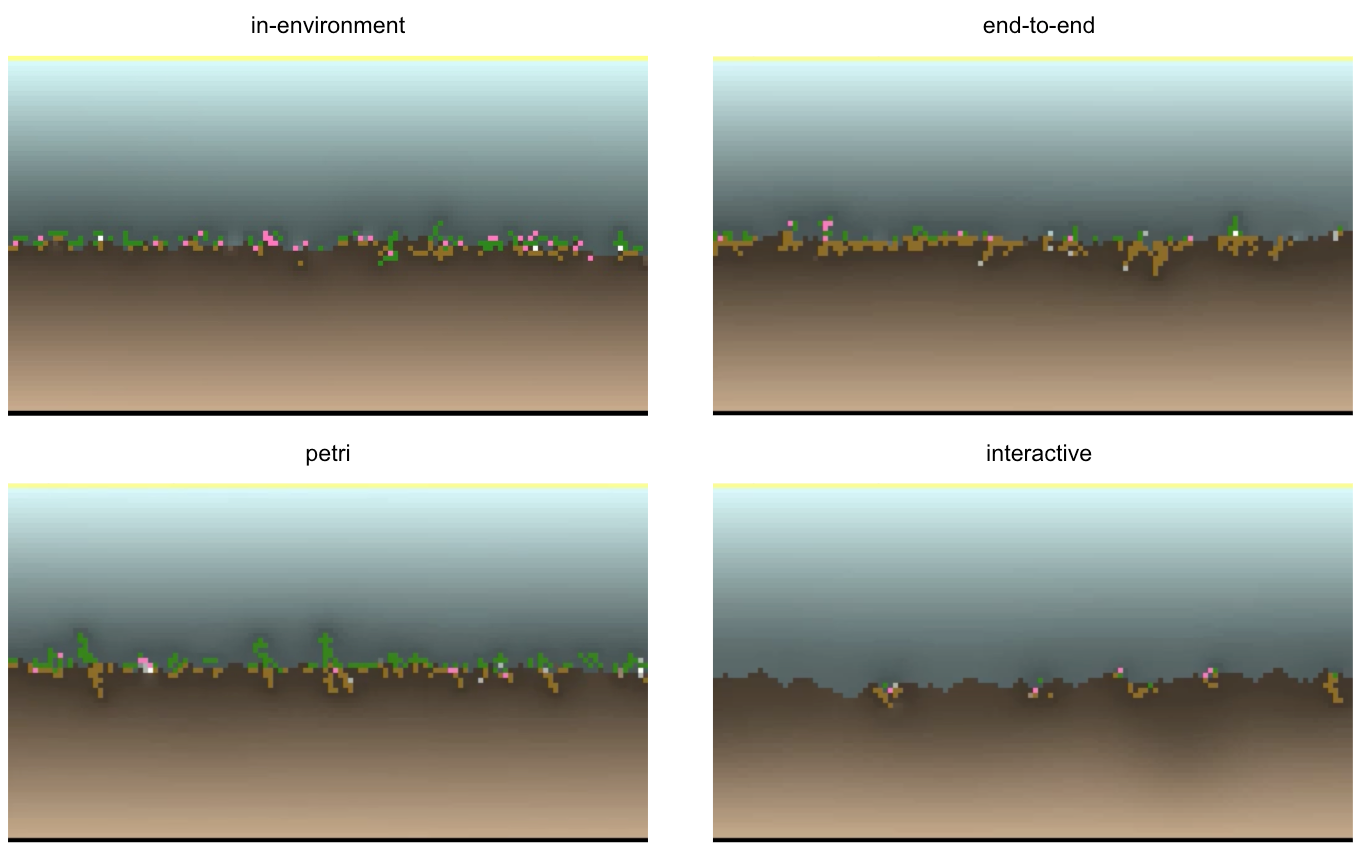}
  \caption{Comparisons of the state of different models evolved in "pestilence" after 1000 steps. All configurations use an extended agent logic with a basic mutator.}
  \label{fig:evolved_comparisons}
\end{figure}

\begin{figure*}
$$
\begin{array}{ c | c | c | c }
\text{initialization} & \text{mutator} & \text{total agents} & \text{extinction \%} \\
\hline
 \text{default} & \text{basic} & 171625 \pm 65775 & 43.75 \\ 
 \text{extracted} & \text{basic} & 210994 \pm 31041 & 6.25 \\
 \text{meta-evolved e2e} & \text{basic} & 250387 \pm 4935 & 0 \\
 \text{meta-evolved petri}& \text{basic} & 216552 \pm 4233 & 0 \\
 \text{meta-evolved petri}& \text{adaptive} & 216368 \pm 4724 & 0 \\
 \text{interactive}& \text{basic} & 216769 \pm 5424 & 0
\end{array}
$$
  \caption{Evaluations of different pairs of (initialization, mutator) run for 1000 steps on pestilence. All pairs use the extended agent logic. Initialization stands for the kind of parameters the initial seed has. default: randomly initialized agent parameter. extracted: parameters of random agents that had survived for 6200 steps in a previous run. meta-evolved e2e: parameters meta-evolved end-to-end. meta-evolved petri: parameters meta-evolved in a petri environment with intercepted reproduction. interactive: parameters evolved with interactive evolution.}
  \label{fig:eval_trained}
\end{figure*}

One approach that we could take to decrease the chance of extinction on pestilence is as follows. We noticed that more than 50\% of runs don't fail within the first 1000 steps. What would happen if we extracted one of the successful agents in one of these runs? Since there is in-environment evolution, it stands to reason that these extracted agents will behave better than our initialized ones. Another way of seeing this experiment is that we are evaluating how well in-environment evolution works. So, we took a random run where after 6200 steps agents were alive, extracted 16 random and unique agent parameters from the surviving population and evaluated them.

Figure~\ref{fig:eval_trained} shows significant improvement over the random initialization that we started with: agents takeover a bigger portion of the environment and only one out of 16 parameters resulted in an extinction event. Figure~\ref{fig:evolved_comparisons} (top left) shows the state of one of the runs after 1000 steps.

These results are encouraging. The biome evolved by itself to become more fit, with the selective pressure exclusively coming from the minimal criterion of triggering a successful reproduce operation through a flower. The next question is: how much could we accomplish if we \textit{did} optimize for our metrics of interest?

\subsection{End-to-end meta-evolution}

We have a vectorizable function that extracts metrics that we want to optimize. From there there are many ways to generate a fitness function by combining these two metrics. Here we choose to create a fitness function that takes as input the total number of agents $a$ and a binary integer $e$ representing whether all agents died by the 1000th step:

$$
\text{fitness}(a, e) = a - e*\texttt{death\_penalty}
$$
Where we chose \texttt{death\_penalty} to be a very high number (1 million).

Now, starting from our initial parameters, any evolutionary strategy can maximize this fitness function. Figure~\ref{fig:meta_evolution_diagram} describes how we perform meta-evolution. Since we are using JAX, we chose to use the PGPE algorithm \citep{Sehnke2008-yi} implemented by the EvoJAX library \citep{evojax2022}. Due to the high cost of this simulation, we chose a population size of 32 and run 30 outer steps of meta-evolution.

Doing this is remarkably easy with Biomaker CA and JAX support, but under the hoods we are, at each outer step of meta-evolution, performing 32 different runs of 1000 steps each, extracting a fitness score and then updating our parameters, which is very complex and costly. So costly that we couldn't run the same algorithm if we used our adaptive mutator due to the doubling in size of parameters.

The results, however, are very positive. Figure~\ref{fig:eval_trained} shows a much higher number of total agents and not a single extinction event, after only 30 steps of meta-evolution. Figure~\ref{fig:evolved_comparisons} (top right)  shows the state of the evolved initialization after 1000 steps.

\subsection{Petri dish meta-evolution}

In this paper we will not do quantitatively better than what we have observed with the end-to-end meta-evolution. Still, that approach was very costly and not well scalable. Moreover, it required to simulate hundreds if not thousands of different agents and their different parameters, while we were only modifying \textit{one} single set of parameters as a result, which appears wasteful.

In this section we explore a much more sample efficient approach for evolving some initial parameters. The core idea is to extract the initial seed and place it into a smaller, "fake" environment. We call these "Petri" environments due to the famous Petri dish experimentations. In this Petri environment, we will only care to observe how the single initial plant behaves and evolve it through some metrics. Afterwards, we will replace this plant's evolved parameters into a "real" environment and see how the new plant behaves. For instance, we will aim to create a plant that strives to be composed of exactly 50 agent cells. This fitness is vastly cheaper to compute than the previous approach: the environment is much smaller, and we know that an agent dies within 300 steps, so we can run the simulation for only that many steps. The target of having exactly 50 agent cells at any time is also easily implementable. The reason why we choose 50 agents and not any other number is partly random and partly because we know that small plants seem to survive more easily in this configuration.

Doing meta-evolution like this would result in evolving a biome that is formed of roughly 50 cells, however there would be nothing that would ensure that: 1) it would still be fertile (it would be capable of reproducing successfully); 2) the 50 cells all come from \textit{one} plant, and not a plant and its offspring generated by a reproduce operation. To solve this, we \textit{intercept} the reproduction operations happening in the environment and do not generate any new seed. This makes sure that the agent remains alone in the environment. While doing so, we still destroy the flower and check how many remaining nutrients the hypothetical new seed would have had. If we believe that the amount of nutrients is sufficient for the seed to be able to survive (chosen with a user-designed threshold), we record this as a "successful reproduction" and continue the simulation. This way, for the entire run the agent will be alone but will try to reproduce. At the end of the run we will have a number estimating the amount of successful reproductions that it would have had if it were in a real environment. We can construct a fitness function by combining the two metrics of interest (number of agents must be 50, and maximize the number of successful reproductions). We refer to the code base for more details on the fitness function.

\begin{figure}[h]
  \includegraphics[width=\columnwidth]{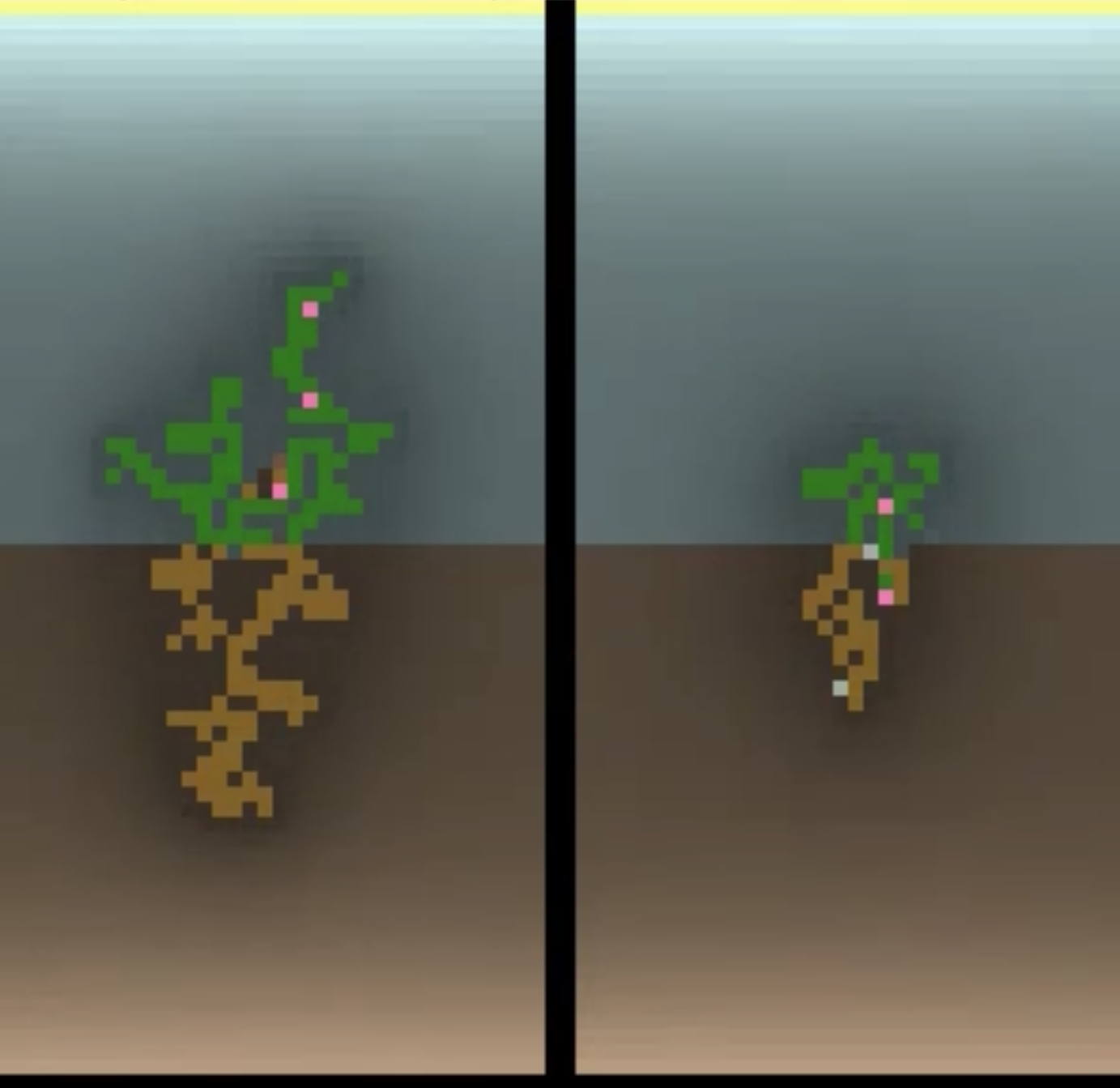}
  \caption{Snapshot after 150 steps in the Petri environment. Left: using the original parameters. Right: using the meta-evolved parameters.}
  \label{fig:petri_intercept_snapshot}
\end{figure}

We perform meta evolution (Figure~\ref{fig:meta_evolution_diagram}) on this Petri environment for 50 outer steps, using PGPE and a population size of 64. Figure~\ref{fig:petri_intercept_snapshot} shows the status of the original parameters (left) and the meta-evolved parameters (right) after 150 steps, right before they start aging. Remember that reproduction is intercepted here; that is why there is only one plant for each slice.

Now we can move the evolved parameters back to the real environment. But what mutator should we use? After all, we have not used any mutator in the inner steps of meta-evolution, so why would any mutator work out of the box once we deploy the model? The best way to find out whether our mutators will work is to try and see what happens. Figure~\ref{fig:eval_trained} shows that both mutators seem to work: while being worse than its end-to-end counterpart based on the number of agents they generated (which we were not trying to maximize with this fitness), they never get extinct.  Figure~\ref{fig:evolved_comparisons} (bottom left) shows a run with the basic mutator.

\subsubsection{The problem of adding mutators in the loop}
The fact that these mutators work out-of-the-box is likely because we chose these mutators to perform tame mutations at initialization. It is very likely that different agent logic and mutator pairs may require coevolution to work well together.

However, doing this effectively is much more complex than one might expect and it requires future research to be solved. Here, we will give some insight as to why this problem is not easily solvable. A well coevolved agent logic and mutator pair must have this property: they have to be \textit{recursively fertile}. By this we mean that not only the original agent needs to be fertile (capable of growing and reproducing), but the same must be true for all of its descendants. In \cite{randazzo2021sr}, the authors show how just ensuring that the parent and a random child are fertile does not ensure at all that their descendants will be. This is true also for longer chains of descendants. In that paper, they solve this problem by creating basins of attraction in the parameter space and show how this appears to create a small amount of variability as well. However, besides there not being any insurance that adding a larger amount of variability would not cause some descendants to be infertile, it also requires to modify the architecture significantly.

We expect that this problem is solvable, but it will require to tweak the agent logic in nontrivial ways and to design a more complex meta-evolution loop than what we have described in this paper.

\subsection{Interactive evolution}

\begin{figure}[h]
  \includegraphics[width=\columnwidth]{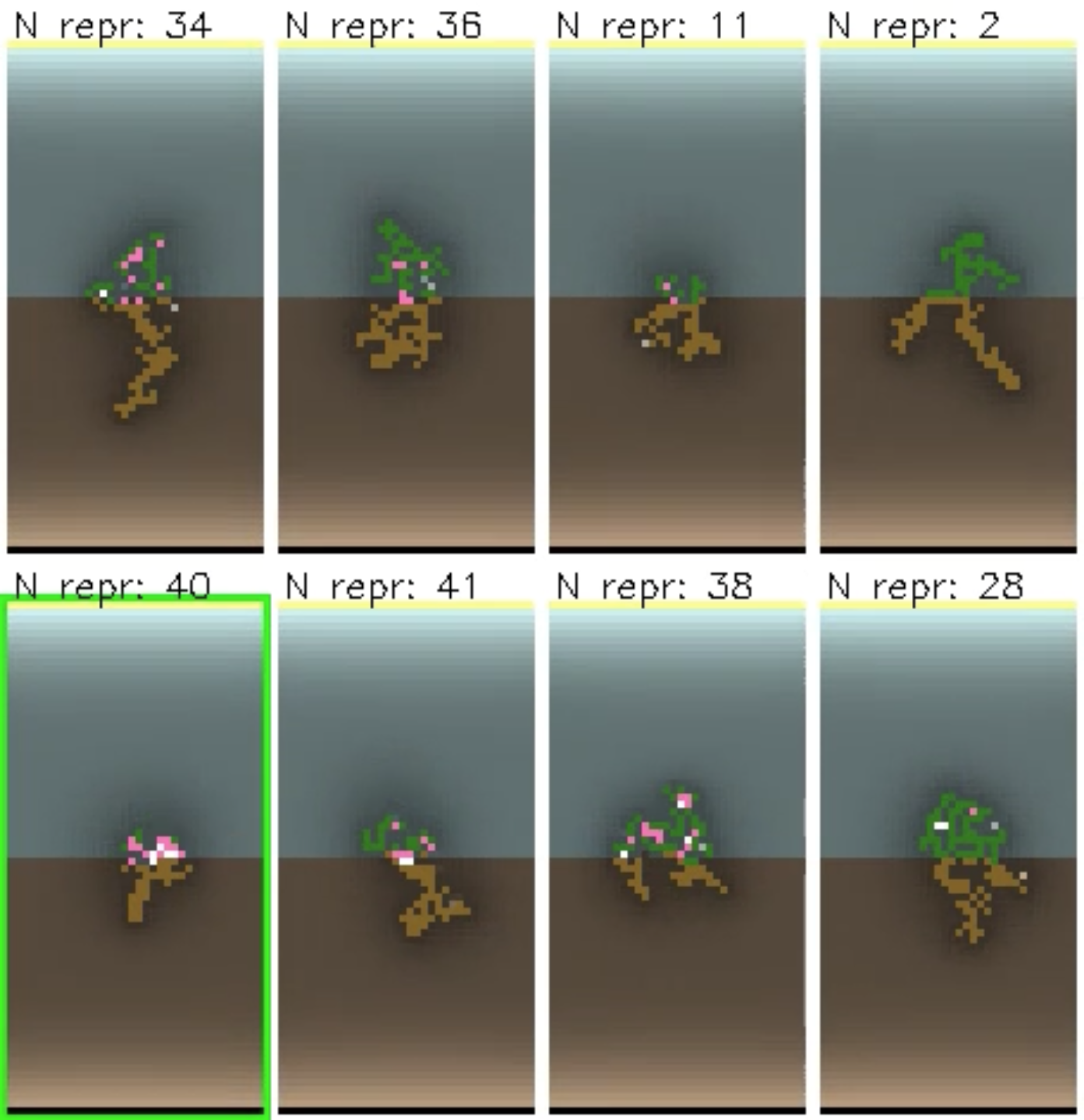}
  \caption{Interface of the interactive evolution. 'N repr' represents the number of successful reproductions that were intercepted. The green rectangle is an addition to indicate what choice was made.}
  \label{fig:picbreeder_example}
\end{figure}

In the final experiment of this paper, we will look at how we can put the human in the loop to evolve a model however they see fit. This experiment is directly inspired by Picbreeder \citep{Secretan2011-sl}, where the authors show how users can interactively evolve some images by repeatedly choosing their favourite offspring out of several options. This approach is strikingly different from traditional optimization approaches because it allows the user to optimize for very vague metrics that could be impossible to define formally, and lets the user maximize for their interests.

This experiment will do the same, but instead of optimizing for an image, the user will have to choose their favourite organism (composed of agent logic and mutator parameters). An image would not be sufficient to show the whole picture, so we approximate the organism representation by showing a video of their lifetime in a Petri environment, and show how many successful reproductions it would have accomplished in real environments. Figure~\ref{fig:picbreeder_example} shows a snapshot of the interface, where we highlight in green what the user may hypothetically choose.

Besides allowing for the user to design their own organisms, we believe that interactive evolution is invaluable for understanding how well an agent logic and a mutator work together. It is in fact common that some evolutionary strategies result in no change at all, or catastrophic mutations being too common, or simply mutations may result in not interesting enough variations. We believe that having a visual feedback of such behaviors may be used to understand what mutators do, and it may lead researchers to choose more effective mutation strategies in search for visible variation at every step.

\begin{figure}[h]
  \includegraphics[width=\columnwidth]{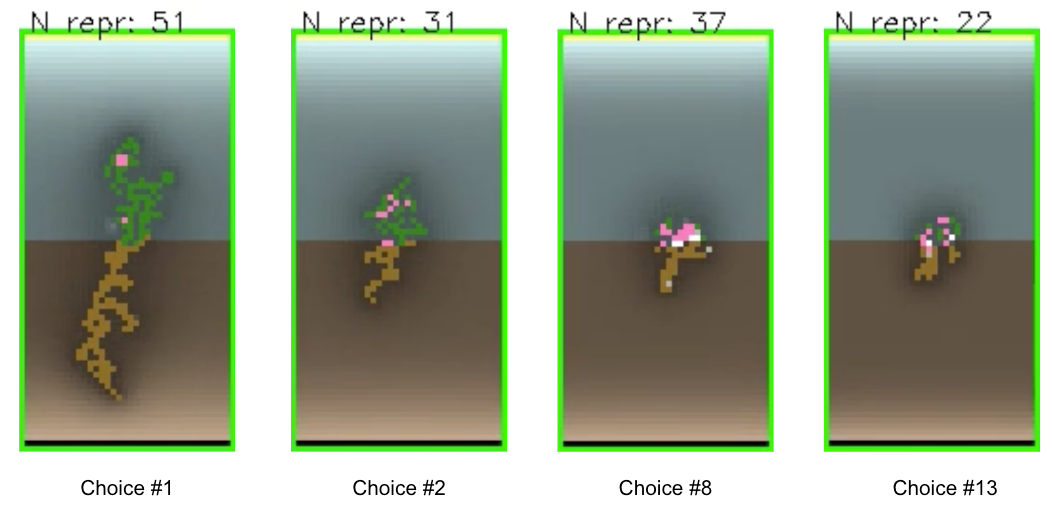}
  \caption{Sequence of choices that were made for the example in this paper for interactive evolution. We only show 4 choices; choice \#13 was the final one.}
  \label{fig:picbreeder_sequence}
\end{figure}

Figure~\ref{fig:picbreeder_sequence} shows an example run that we performed to showcase the possibilities of interactive evolution. Until our eight choice, we were just choosing semi-random plants, but the eight one piqued our interest (this choice is also visible in Figure~\ref{fig:picbreeder_example}). While an image cannot justify it entirely, the behavior we observed was a plant that grew and later exploded into many flowers before dying. We then tried to maximize for this behavior until our thirteenth choice, where we stopped. We then deployed the evolved plant into a larger, real environment and Figure~\ref{fig:eval_trained} shows how, somewhat surprisingly, this plant behaves extremely well and does not ever get extinct.

Figure~\ref{fig:evolved_comparisons} (bottom right) shows a snapshot of its behavior at 1000 steps, but it doesn't really do it justice. We observed that the first few offspring of the original plant all behave similarly to it and create flowers in an explosive behavior. However, this property very quickly disappears down the descendant line. This is potentially a problem, and it certainly shows that we don't have much control on how we plan to evolve our organisms. It is well known that in biology some original morphogenetic decisions become fixed and evolution doesn't vary them anymore. Our models don't have this property, and it would both be interesting to be able to design some organisms that would forever store the property of exploding into flowers in their dna, and we suspect this capacity to be essential to improve in our quest for complexification.

\section{Discussion}

Biomaker is designed to be a playground for researchers and tech savvies that have interest in Artificial Life, complexification, open endedness and ecology. We released all the code and showed a selection of configurations and experiment in order to inspire the reader to contribute, build on top of Biomaker CA, and most importantly improve on the state of the art on the topics of their choice.

While we expect that every researcher would find something different to explore with Biomaker CA, we will list some questions that piqued our interest in the process of writing this paper. What are the laws of physics that stand at the edge of chaos? What universes allow for interesting interactions to occur, and which are boring? What would be the effect of adding new materials and agent operations? What are the best architectures/basic building blocks for creating complex morphogenetic organisms? What mutators work best alongside certain agent logics for evolving quickly into something interesting and potentially arbitrarily complex? Is asexual reproduction the best way, or should we create a reproduce operation that is sexual? How can we design organisms that evolve into having unchanging key properties that, while they will still evolve in different ways, won't ever lose such properties? How can we create recursively fertile agents without having to simulate entire environments? How can the human be in the loop as much as possible, or as little as possible? How can we design complex biomes that don't have abrupt extinction events? Can we understand ecology in the real world by simulating biomes in Biomaker CA, or variants of it?

Finally, one of our desires is that this framework will allow to discover successful and reusable configurations, architectures, mutator strategies \textit{and parameters} so that the community can start building on top of each other to evolve increasingly complex lifeforms not only in isolated research projects, but also \textit{across} projects.

While the number of open questions that seem to be possible to be answered with Biomaker CA is vast, Biomaker CA has certainly many limitations that makes it unsuitable for answering \textit{all} question in ALife, complexification theory and ecology. The most evident limitation is the incapacity of agents to move. Even if adding a 'move' operation is trivial, CA-based worlds are not friendly to this kind of movements, since the moving of one cell would leave the rest of the organism behind. Therefore, Biomaker CA is chiefily focused on plant-like agents and at most very simple moving organisms. Also, space and time are discrete, and space is a 2d grid, which bring several limitations and unrealistic parallels with the real world. Also compositionality of organoids is not well supported: we can certainly create complex organisms with Biomaker CA, but we can't make \textit{containers} that compose higher and higher levels of complexity easily. For all of these restrictions, we expect that eventually other frameworks may come to provide more or different possibilities. However, we believe that it may be detrimental to add further complexity in out frameworks until we solved some of the open questions that seem feasible to tackle with Biomaker CA.

\footnotesize
\bibliographystyle{apalike}
\bibliography{example} 

\begin{thebibliography}{}

\bibitem[Bittker, 2019]{bittkersandspiel}
Bittker, M. (30.04.2019).
\newblock Making sandspiel.

\bibitem[Bradbury et~al., 2018]{jax2018github}
Bradbury, J., Frostig, R., Hawkins, P., Johnson, M.~J., Leary, C., Maclaurin,
  D., Necula, G., Paszke, A., Vander{P}las, J., Wanderman-{M}ilne, S., and
  Zhang, Q. (2018).
\newblock {JAX}: composable transformations of {P}ython+{N}um{P}y programs.

\bibitem[Brant and Stanley, 2017]{brant2017as}
Brant, J.~C. and Stanley, K.~O. (2017).
\newblock Minimal criterion coevolution: A new approach to open-ended search.
\newblock In {\em Proceedings of the Genetic and Evolutionary Computation
  Conference}, GECCO '17, page 67–74, New York, NY, USA. Association for
  Computing Machinery.

\bibitem[Cavuoti et~al., 2022]{cavuoti2022}
Cavuoti, L., Sacco, F., Randazzo, E., and Levin, M. (2022).
\newblock {Adversarial Takeover of Neural Cellular Automata}.
\newblock volume ALIFE 2022: The 2022 Conference on Artificial Life of {\em
  ALIFE 2022: The 2022 Conference on Artificial Life}.
\newblock 38.

\bibitem[Chan, 2019]{chan2019lenia}
Chan, B. W.-C. (2019).
\newblock Lenia: Biology of artificial life.
\newblock {\em Complex Systems}, 28(3):251–286.

\bibitem[Chan, 2023]{chan2023largescale}
Chan, B. W.-C. (2023).
\newblock Towards large-scale simulations of open-ended evolution in continuous
  cellular automata.

\bibitem[Channon, 2006]{channon2006geb}
Channon, A. (2006).
\newblock Unbounded evolutionary dynamics in a system of agents that actively
  process and transform their environment.
\newblock {\em Genetic Programming and Evolvable Machines}, 7:253--281.

\bibitem[Clune et~al., 2008]{Clune2008-zs}
Clune, J., Misevic, D., Ofria, C., Lenski, R.~E., Elena, S.~F., and
  Sanju{\'a}n, R. (2008).
\newblock Natural selection fails to optimize mutation rates for long-term
  adaptation on rugged fitness landscapes.
\newblock {\em PLoS Comput. Biol.}, 4(9):e1000187.

\bibitem[Clune et~al., 2013]{Clune2013-yk}
Clune, J., Mouret, J.-B., and Lipson, H. (2013).
\newblock The evolutionary origins of modularity.
\newblock {\em Proc. Biol. Sci.}, 280(1755):20122863.

\bibitem[Ebner et~al., 2001]{ebner2001}
Ebner, M., Shackleton, M., and Shipman, R. (2001).
\newblock How neutral networks influence evolvability.
\newblock {\em Complexity}, 7(2):19--33.

\bibitem[Forestier et~al., 2017]{forestier2017intrinsically}
Forestier, S., Portelas, R., Mollard, Y., and Oudeyer, P.-Y. (2017).
\newblock Intrinsically motivated goal exploration processes with automatic
  curriculum learning.

\bibitem[Frans and Isola, 2022]{frans2022powderworld}
Frans, K. and Isola, P. (2022).
\newblock Powderworld: A platform for understanding generalization via rich
  task distributions.

\bibitem[Gajewski et~al., 2019]{gajewski2019}
Gajewski, A., Clune, J., Stanley, K.~O., and Lehman, J. (2019).
\newblock Evolvability es: Scalable and direct optimization of evolvability.
\newblock GECCO '19, page 107–115, New York, NY, USA. Association for
  Computing Machinery.

\bibitem[Gardner, 1970]{10.2307/24927642}
Gardner, M. (1970).
\newblock Mathematical games.
\newblock {\em Scientific American}, 223(4):120--123.

\bibitem[Glickman and Sycara, 2000]{glickman2000}
Glickman, M. and Sycara, K. (2000).
\newblock Reasons for premature convergence of self-adapting mutation rates.
\newblock In {\em Proceedings of the 2000 Congress on Evolutionary Computation.
  CEC00 (Cat. No.00TH8512)}, volume~1, pages 62--69 vol.1.

\bibitem[Grefenstette, 1999]{grefenstette1999}
Grefenstette, J. (1999).
\newblock Evolvability in dynamic fitness landscapes: a genetic algorithm
  approach.
\newblock In {\em Proceedings of the 1999 Congress on Evolutionary
  Computation-CEC99 (Cat. No. 99TH8406)}, volume~3, pages 2031--2038 Vol. 3.

\bibitem[Heinemann, 2023]{heinemann2023}
Heinemann, C. (2023).
\newblock Alien project.

\bibitem[Kashtan and Alon, 2005]{Kashtan2005-ym}
Kashtan, N. and Alon, U. (2005).
\newblock Spontaneous evolution of modularity and network motifs.
\newblock {\em Proc. Natl. Acad. Sci. U. S. A.}, 102(39):13773--13778.

\bibitem[Kirschner and Gerhart, 1998]{kirschner1998}
Kirschner, M. and Gerhart, J. (1998).
\newblock Evolvability.
\newblock {\em Proceedings of the National Academy of Sciences},
  95(15):8420--8427.

\bibitem[Kramer, 2010]{Kramer2010}
Kramer, O. (2010).
\newblock Evolutionary self-adaptation: a survey of operators and strategy
  parameters.
\newblock {\em Evolutionary Intelligence}, 3(2):51--65.

\bibitem[Lehman and Miikkulainen, 2015]{lehman2015ex}
Lehman, J. and Miikkulainen, R. (2015).
\newblock Enhancing divergent search through extinction events.
\newblock In {\em Proceedings of the 2015 Annual Conference on Genetic and
  Evolutionary Computation}, GECCO '15, page 951–958, New York, NY, USA.
  Association for Computing Machinery.

\bibitem[Lehman and Stanley, 2008]{Lehman2008ExploitingOT}
Lehman, J. and Stanley, K.~O. (2008).
\newblock Exploiting open-endedness to solve problems through the search for
  novelty.
\newblock In {\em IEEE Symposium on Artificial Life}.

\bibitem[Lehman and Stanley, 2011]{lehman2011as}
Lehman, J. and Stanley, K.~O. (2011).
\newblock Improving evolvability through novelty search and self-adaptation.
\newblock In {\em 2011 IEEE Congress of Evolutionary Computation (CEC)}, pages
  2693--2700.

\bibitem[Liu et~al., 2023]{liu2023summary}
Liu, Y., Han, T., Ma, S., Zhang, J., Yang, Y., Tian, J., He, H., Li, A., He,
  M., Liu, Z., Wu, Z., Zhu, D., Li, X., Qiang, N., Shen, D., Liu, T., and Ge,
  B. (2023).
\newblock Summary of chatgpt/gpt-4 research and perspective towards the future
  of large language models.

\bibitem[Mengistu et~al., 2016]{mengistu2016}
Mengistu, H., Lehman, J., and Clune, J. (2016).
\newblock Evolvability search: Directly selecting for evolvability in order to
  study and produce it.
\newblock In {\em Proceedings of the Genetic and Evolutionary Computation
  Conference 2016}, GECCO '16, page 141–148, New York, NY, USA. Association
  for Computing Machinery.

\bibitem[Miconi, 2008]{miconi2008}
Miconi, T. (2008).
\newblock Evosphere: Evolutionary dynamics in a population of fighting virtual
  creatures.
\newblock In {\em 2008 IEEE Congress on Evolutionary Computation (IEEE World
  Congress on Computational Intelligence)}, pages 3066--3073.

\bibitem[Miconi and Channon, 2005]{miconi2005tr}
Miconi, T. and Channon, A. (2005).
\newblock A virtual creatures model for studies in artificial evolution.
\newblock In {\em 2005 IEEE Congress on Evolutionary Computation}, volume~1,
  pages 565--572 Vol.1.

\bibitem[Mordvintsev et~al., 2022a]{mordvintsev2022particle}
Mordvintsev, A., Niklasson, E., and Randazzo, E. (2022a).
\newblock Particle lenia and the energy-based formulation.

\bibitem[Mordvintsev et~al., 2022b]{mordvintsev2022iso}
Mordvintsev, A., Randazzo, E., and Fouts, C. (2022b).
\newblock {Growing Isotropic Neural Cellular Automata}.
\newblock volume ALIFE 2022: The 2022 Conference on Artificial Life of {\em
  ALIFE 2022: The 2022 Conference on Artificial Life}.
\newblock 65.

\bibitem[Mordvintsev et~al., 2020]{mordvintsev2020growing}
Mordvintsev, A., Randazzo, E., Niklasson, E., and Levin, M. (2020).
\newblock Growing neural cellular automata.
\newblock {\em Distill}.
\newblock https://distill.pub/2020/growing-ca.

\bibitem[Mouret and Clune, 2015]{Mouret2015IlluminatingSS}
Mouret, J.-B. and Clune, J. (2015).
\newblock Illuminating search spaces by mapping elites.
\newblock {\em ArXiv}, abs/1504.04909.

\bibitem[Neumann and Burks, 1966]{neumann1966ca}
Neumann, J.~V. and Burks, A.~W. (1966).
\newblock {\em Theory of Self-Reproducing Automata}.
\newblock University of Illinois Press, USA.

\bibitem[{Nolla Games}, 2019]{noita2019}
{Nolla Games} (2019).
\newblock Noita.

\bibitem[Ofria and Wilke, 2004]{Ofria2004-zg}
Ofria, C. and Wilke, C.~O. (2004).
\newblock Avida: a software platform for research in computational evolutionary
  biology.
\newblock {\em Artif. Life}, 10(2):191--229.

\bibitem[Pigliucci, 2008]{Pigliucci2008-vx}
Pigliucci, M. (2008).
\newblock Is evolvability evolvable?
\newblock {\em Nat. Rev. Genet.}, 9(1):75--82.

\bibitem[Plantec et~al., 2022]{plantec2022flowlenia}
Plantec, E., Hamon, G., Etcheverry, M., Oudeyer, P.-Y., Moulin-Frier, C., and
  Chan, B. W.-C. (2022).
\newblock Flow-lenia: Towards open-ended evolution in cellular automata through
  mass conservation and parameter localization.

\bibitem[Plested and Gedeon, 2022]{plested2022deep}
Plested, J. and Gedeon, T. (2022).
\newblock Deep transfer learning for image classification: a survey.

\bibitem[Pugh et~al., 2016]{Pugh2016-ao}
Pugh, J.~K., Soros, L.~B., and Stanley, K.~O. (2016).
\newblock Quality diversity: A new frontier for evolutionary computation.
\newblock {\em Frontiers in Robotics and AI}, 3.

\bibitem[Randazzo et~al., 2023]{r2023growing}
Randazzo, E., Mordvintsev, A., and Fouts, C. (2023).
\newblock Growing steerable neural cellular automata.

\bibitem[Randazzo et~al., 2021a]{randazzo2021adversarial}
Randazzo, E., Mordvintsev, A., Niklasson, E., and Levin, M. (2021a).
\newblock Adversarial reprogramming of neural cellular automata.
\newblock {\em Distill}.
\newblock https://distill.pub/selforg/2021/adversarial.

\bibitem[Randazzo et~al., 2020]{randazzo2020self-classifying}
Randazzo, E., Mordvintsev, A., Niklasson, E., Levin, M., and Greydanus, S.
  (2020).
\newblock Self-classifying mnist digits.
\newblock {\em Distill}.
\newblock https://distill.pub/2020/selforg/mnist.

\bibitem[Randazzo et~al., 2021b]{randazzo2021sr}
Randazzo, E., Versari, L., and Mordvintsev, A. (2021b).
\newblock {Recursively Fertile Self-replicating Neural Agents}.
\newblock volume ALIFE 2021: The 2021 Conference on Artificial Life of {\em
  Artificial Life Conference Proceedings}, page~58.

\bibitem[Ray, 1991]{ray-approach-to-synthesis-1991}
Ray, T.~S. (1991).
\newblock An approach to the synthesis of life.
\newblock In {C. Langton, C. Taylor, J. D. Farmer, S. Rasmussen}, editor, {\em
  Artificial Life II, Santa Fe Institute Studies in the Sciences of
  Complexity}, volume vol. XI, pages 371--408. Addison-Wesley, Redwood City,
  CA.

\bibitem[Reisinger and Miikkulainen, 2006]{Reisinger2006-xi}
Reisinger, J. and Miikkulainen, R. (2006).
\newblock Selecting for evolvable representations.
\newblock In {\em Proceedings of the 8th annual conference on Genetic and
  evolutionary computation}, New York, NY, USA. ACM.

\bibitem[Robinson, 2023]{robinsonlife}
Robinson, M. (2023).
\newblock The life engine.

\bibitem[Schmickl et~al., 2016]{Schmickl2016-nc}
Schmickl, T., Stefanec, M., and Crailsheim, K. (2016).
\newblock How a life-like system emerges from a simple particle motion law.
\newblock {\em Sci. Rep.}, 6:37969.

\bibitem[Schmidhuber, 2013]{schmidhuber2013}
Schmidhuber, J. (2013).
\newblock Powerplay: Training an increasingly general problem solver by
  continually searching for the simplest still unsolvable problem.
\newblock {\em Frontiers in Psychology}, 4.

\bibitem[Secretan et~al., 2011]{Secretan2011-sl}
Secretan, J., Beato, N., D'Ambrosio, D.~B., Rodriguez, A., Campbell, A.,
  Folsom-Kovarik, J.~T., and Stanley, K.~O. (2011).
\newblock Picbreeder: a case study in collaborative evolutionary exploration of
  design space.
\newblock {\em Evol. Comput.}, 19(3):373--403.

\bibitem[Sehnke et~al., 2008]{Sehnke2008-yi}
Sehnke, F., Osendorfer, C., R{\"u}ckstie{\ss}, T., Graves, A., Peters, J., and
  Schmidhuber, J. (2008).
\newblock Policy gradients with {Parameter-Based} exploration for control.
\newblock In {\em Artificial Neural Networks - {ICANN} 2008}, pages 387--396.
  Springer Berlin Heidelberg.

\bibitem[Sims, 1994]{sims1994}
Sims, K. (1994).
\newblock Evolving 3d morphology and behavior by competition.
\newblock {\em Artificial Life}, 1(4):353--372.

\bibitem[Sinapayen, 2023]{Sinapayen2023}
Sinapayen, L. (2023).
\newblock Self-replication, spontaneous mutations, and exponential genetic
  drift in neural cellular automata.
\newblock {\em Qeios}.

\bibitem[Soros and Stanley, 2014]{soros2014}
Soros, L. and Stanley, K. (2014).
\newblock Identifying necessary conditions for open-ended evolution through the
  artificial life world of chromaria.
\newblock volume ALIFE 14: The Fourteenth International Conference on the
  Synthesis and Simulation of Living Systems of {\em Artificial Life Conference
  Proceedings}, pages 793--800.

\bibitem[Spector et~al., 2007]{spector2007}
Spector, L., Klein, J., and Feinstein, M. (2007).
\newblock Division blocks and the open-ended evolution of development, form,
  and behavior.
\newblock In {\em Proceedings of the 9th Annual Conference on Genetic and
  Evolutionary Computation}, GECCO '07, page 316–323, New York, NY, USA.
  Association for Computing Machinery.

\bibitem[Suarez et~al., 2019]{suarez2019neural}
Suarez, J., Du, Y., Isola, P., and Mordatch, I. (2019).
\newblock Neural mmo: A massively multiagent game environment for training and
  evaluating intelligent agents.

\bibitem[Tang et~al., 2022]{evojax2022}
Tang, Y., Tian, Y., and Ha, D. (2022).
\newblock Evojax.
\newblock {\em Proceedings of the Genetic and Evolutionary Computation
  Conference Companion}.

\bibitem[Team et~al., 2021]{team2021openended}
Team, O. E.~L., Stooke, A., Mahajan, A., Barros, C., Deck, C., Bauer, J.,
  Sygnowski, J., Trebacz, M., Jaderberg, M., Mathieu, M., McAleese, N.,
  Bradley-Schmieg, N., Wong, N., Porcel, N., Raileanu, R., Hughes-Fitt, S.,
  Dalibard, V., and Czarnecki, W.~M. (2021).
\newblock Open-ended learning leads to generally capable agents.

\bibitem[Ventrella, 1998]{ventrella1998genepool}
Ventrella, J. (1998).
\newblock Attractiveness vs. efficiency (how mate preference affects location
  in the evolution of artificial swimming organisms).
\newblock In {\em Proceedings of the Sixth International Conference on
  Artificial Life}, ALIFE, page 178–186, Cambridge, MA, USA. MIT Press.

\bibitem[Ventrella, 2017]{ventrella2023}
Ventrella, J. (2017).
\newblock Clusters.

\bibitem[Wagner and Altenberg, 1996]{Wagner1996-rt}
Wagner, G.~P. and Altenberg, L. (1996).
\newblock Perspective: Complex adaptations and the evolution of evolvability.
\newblock {\em Evolution}, 50(3):967--976.

\bibitem[Wang et~al., 2019]{wang2019paired}
Wang, R., Lehman, J., Clune, J., and Stanley, K.~O. (2019).
\newblock Paired open-ended trailblazer (poet): Endlessly generating
  increasingly complex and diverse learning environments and their solutions.

\bibitem[Wang et~al., 2020]{wang2020enhanced}
Wang, R., Lehman, J., Rawal, A., Zhi, J., Li, Y., Clune, J., and Stanley, K.~O.
  (2020).
\newblock Enhanced poet: Open-ended reinforcement learning through unbounded
  invention of learning challenges and their solutions.

\bibitem[Yaeger, 1995]{yaeger1995}
Yaeger, L. (1995).
\newblock Computational genetics, physiology, metabolism, neural systems,
  learning, vision, and behavior or polyworld: Life in a new context.

\end{thebibliography}

\end{document}